\documentclass{article}

\PassOptionsToPackage{numbers, compress}{natbib}
\usepackage[eandd,final]{neurips_2026}

\usepackage[utf8]{inputenc}
\usepackage[T1]{fontenc}
\usepackage{hyperref}
\usepackage{url}
\usepackage{booktabs}
\usepackage{amsfonts}
\usepackage{amsmath,amssymb}
\usepackage{nicefrac}
\usepackage{microtype}
\usepackage{xcolor}
\usepackage{graphicx}
\usepackage{enumitem}
\usepackage{multirow}
\usepackage{placeins}
\usepackage{algorithm}
\usepackage{algorithmic}
\usepackage{caption}
\usepackage{subcaption}
\usepackage{tcolorbox}
\tcbuselibrary{skins}
\usepackage{tikz}
\usetikzlibrary{arrows.meta,positioning,shapes.geometric}

\hypersetup{colorlinks=true, linkcolor=blue!60!black, citecolor=blue!60!black, urlcolor=blue!60!black}

\title{Solving Every Step Is Not Enough: Milestone Oracles\\Reveal a Composition Gap in LLM Math Reasoning}

\author{%
  Zhuohan Wang$^{1*\dagger\ddagger}$, Haoran Ma$^{1*}$, Tianyu Wu$^{1*}$,\\
  \textbf{Yuanlin Duan$^{2}$, Zichun Liao$^{1}$, Jieming Yu$^{3\dagger}$}\\
  $^{1}$Harvard University \quad $^{2}$Rutgers University\\
  $^{3}$The Hong Kong University of Science and Technology
}

\newif\ifchecklist\checklistfalse

\begin{document}
\maketitle

{\renewcommand{\thefootnote}{}\footnotetext{$^{*}$Equal contribution.}\footnotetext{$^{\dagger}$Corresponding authors: \texttt{zhuohan\_wang@fas.harvard.edu}, \texttt{jyucu@connect.ust.hk}.}\footnotetext{$^{\ddagger}$Project lead.}}
\setcounter{footnote}{0}

\begin{abstract}
Large language models (LLMs) can solve every intermediate step of a multi-step math problem on its own and still fail the full problem, even when given a roadmap of the steps and all of their answers. We introduce OracleLadder, a diagnostic evaluation that locates where LLM math reasoning fails by giving the model increasing levels of oracle help. For each problem, a teacher model writes a fixed roadmap of intermediate sub-goals (milestones), and a deterministic symbolic verifier grades every answer. Testing the model with no help, with the roadmap, with the roadmap plus the milestone answers, and on each milestone alone sorts each failure into one of five reasoning gaps. On 354 NuminaMath problems and six models from 8B to 671B parameters (Qwen3, gpt-oss, Llama 3.3, DeepSeek-V3.1), the largest gap for every model is the composition gap, a stricter form of the compositionality gap. It covers 33--48\% of problems, and 24--37\% after removing problems that an LLM review flags as grading errors. Accuracy and milestone-help recovery rank the two strongest models differently, and two RLVR runs with similar accuracy gains move problems differently. The roadmap effect replicates on MATH500 and AIME 2024/25, per-problem recovery agrees for 83--87\% of problems under an independent second teacher, and the help ladder carries over to code generation. We release the data, roadmaps, prompts, and code at \url{https://github.com/slark-prime/OracleLadder}.
\end{abstract}

\section{Introduction}

Math benchmarks for large language models (LLMs) range from grade-school word problems~\citep{Cobbe2021VerifiersGSM8K} to competition~\citep{hendrycksmath2021} and olympiad or frontier problems~\citep{he2024olympiadbench,glazer2024frontiermath}. Scores on many of them are rising quickly with reasoning-trained models~\citep{deepseek2025r1,deepmind2025alphaproof}. An accuracy number records how often a model was right, but it compresses different failures into one value~\citep{Dehghani2021BenchmarkLottery}. Two models with the same score can fail for different reasons, and a training run that raises the score can change different parts of the reasoning. To compare models or training runs, we need to know which part of the reasoning failed: planning the route, executing an intermediate step, combining intermediate results, or solving a step that is out of reach.

Existing evaluations approach this from two sides. Trace-based methods score the steps a model writes~\citep{Wei2022CoT,lightman2023verify,xia2024reasoneval}. Perturbation methods change the problem itself~\citep{wu2024reasoning,press2023compositionality}. Closer to our setting, CHAMP~\citep{mao2024champ} and \citet{agrawal2024hint} add hints to math problems and measure how much they help. We-Math~\citep{qiao2025wemath} splits visual math problems into sub-problems and checks whether a model that solves the parts also solves the whole. We hold each problem fixed and control exactly what intermediate information the model receives. A teacher writes one roadmap of milestones per problem. We give the model this roadmap, then the roadmap plus every milestone answer, and we separately test each milestone alone. Because LLM judges can over-accept invalid answers~\citep{Zheng2023MTBench}, a deterministic symbolic verifier grades every answer.

This readout matters most for reinforcement learning with verifiable rewards (RLVR)~\citep{shao2024deepseekmath,deepseek2025r1,yu2025dapo}. The reward checks only the final answer, and process-supervised methods that score intermediate steps~\citep{uesato2022solving,lightman2023verify} are still evaluated by final accuracy after training. A higher post-training score therefore leaves open whether training reduced planning failures, step-execution failures, composition failures, or capability gaps. A diagnostic that compares models on the same problems and roadmaps can also track, checkpoint by checkpoint, which problems training moves from one gap to another.

\begin{figure}[t]
\centering
\includegraphics[width=0.92\textwidth]{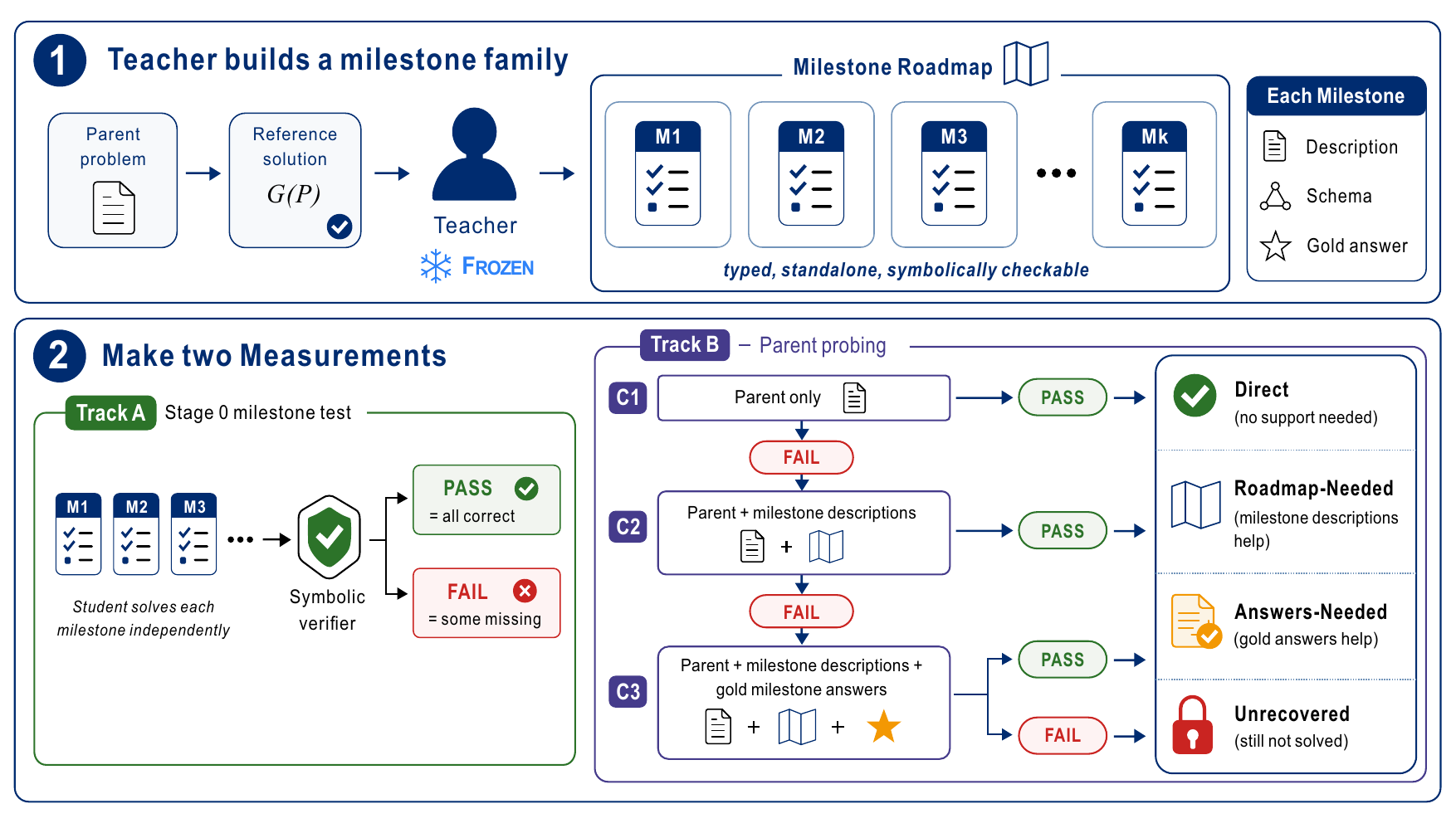}
\caption{\textbf{OracleLadder workflow.} A teacher compiles a fixed milestone roadmap from the parent problem and its reference solution $G(P)$. The student is tested on Track~A (milestone-only) and on Track~B (the parent under help levels $C_1$, $C_2$, $C_3$). Crossing the two tracks gives the reasoning-gap diagnosis.}
\label{fig:protocol_overview}
\end{figure}

\paragraph{OracleLadder.}
For each parent problem, a teacher builds a milestone roadmap: typed, symbolically checkable intermediate sub-goals with gold answers, held fixed across probes and evaluated models. OracleLadder then gives the model three rungs of help on the parent problem. $C_1$ gives no help, $C_2$ gives the roadmap, and $C_3$ gives the roadmap plus the gold milestone answers. Matched controls replace the roadmap with one from a different problem, replace it with a generic decomposition prompt, or pair the correct roadmap with wrong answers. They test whether any lift comes from problem-matched information. A separate milestone test (Stage~0) asks whether the model can solve each milestone alone. Crossing the two measurements separates a missing step from steps that are available but not combined (Figure~\ref{fig:protocol_overview}).

\paragraph{What OracleLadder shows.}
On 354 NuminaMath problems and six models from 8B to 671B parameters, the largest of the five gaps for every model is the \textsc{composition-gap}. The model passes the milestone test, yet no rung of help solves the problem (33--48\% of problems). This holds for the three models from other lineages as well as for the Qwen models whose base model screened the problems. It is a stricter form of the compositionality gap of \citet{press2023compositionality}, because the model fails even with the roadmap and every milestone answer in hand. After removing problems that an LLM rubric review flags as verifier or format errors, the category still covers 24--37\% of problems for every model. Raw outputs show such a failure directly. On one audited problem, Qwen3-8B (the post-trained release, thinking disabled) solves both milestones alone in all eight attempts and receives both answers. It still fails all eight attempts, while gpt-oss-20b solves the problem from the same roadmap and answers in 6 of 8. The readout also separates models that one accuracy number groups together. DeepSeek-V3.1 has the highest per-attempt accuracy, yet gpt-oss-20b recovers more problems with milestone help. Two RLVR runs with similar MATH gains shift problems between help levels in different ways.

\paragraph{The roadmap effect transfers.}
On the 38 AIME 2024/25 problems that Qwen3-8B fails directly, the matched roadmap solves 45\% (pass@8), against 13\% for a wrong-problem or a generic roadmap. On MATH500~\citep{lightman2023verify} it raises per-attempt accuracy from 10.0\% to 37.9\%, while wrong-problem and generic roadmaps stay at 7.9\% and 9.6\%. A 10-problem code-generation pilot on LiveCodeBench~\citep{jain2025livecodebench}, graded by running the tests, shows the same ladder: 6.5\%, 41.2\%, and 88.8\% of attempts pass with no help, with the roadmap, and with the roadmap plus verified helper functions. Per-problem recovery agrees for 83--87\% of problems when an independent teacher writes the roadmaps, and for 85\% when the same teacher writes roadmaps four times finer. Re-screening the set with each panel model as the anchor keeps the model ranking (mean Kendall $\tau{=}0.80$, all 15 anchor pairs positive).

\paragraph{Contributions.}
\begin{enumerate}[nosep,leftmargin=*]
  \item We introduce OracleLadder, a symbolically verified diagnostic that locates where LLM math reasoning fails by crossing a milestone-alone test with a ladder of oracle help on the full problem.
  \item We find that the \textsc{composition-gap} is the largest of five reasoning gaps for all six models (33--48\% of problems, 24--37\% after removing verifier and format errors), and that the readout separates models and RLVR runs that aggregate accuracy groups together.
  \item We show that the roadmap effect replicates on MATH500 and AIME 2024/25 and that the same ladder appears in a code-generation pilot. Per-problem recovery stays stable when a second teacher writes the roadmaps or the roadmaps become four times finer, and model rankings stay stable under different screening models.
  \item We release the 354-problem diagnostic set, roadmaps, prompts, verifier, matched controls, per-model results, audit labels, and the replication suites.
\end{enumerate}
\section{Method}
\label{sec:method}

\definecolor{wexParent}{HTML}{475569}
\definecolor{wexMile}{HTML}{3B82F6}
\definecolor{wexAnswer}{HTML}{8B5CF6}
\definecolor{wexUnrec}{HTML}{9CA3AF}
\definecolor{wexCheck}{HTML}{059669}
\definecolor{wexFail}{HTML}{DC2626}
\definecolor{wexPass}{HTML}{059669}
\definecolor{wexNeutral}{HTML}{6B7280}

\begin{figure}[!ht]
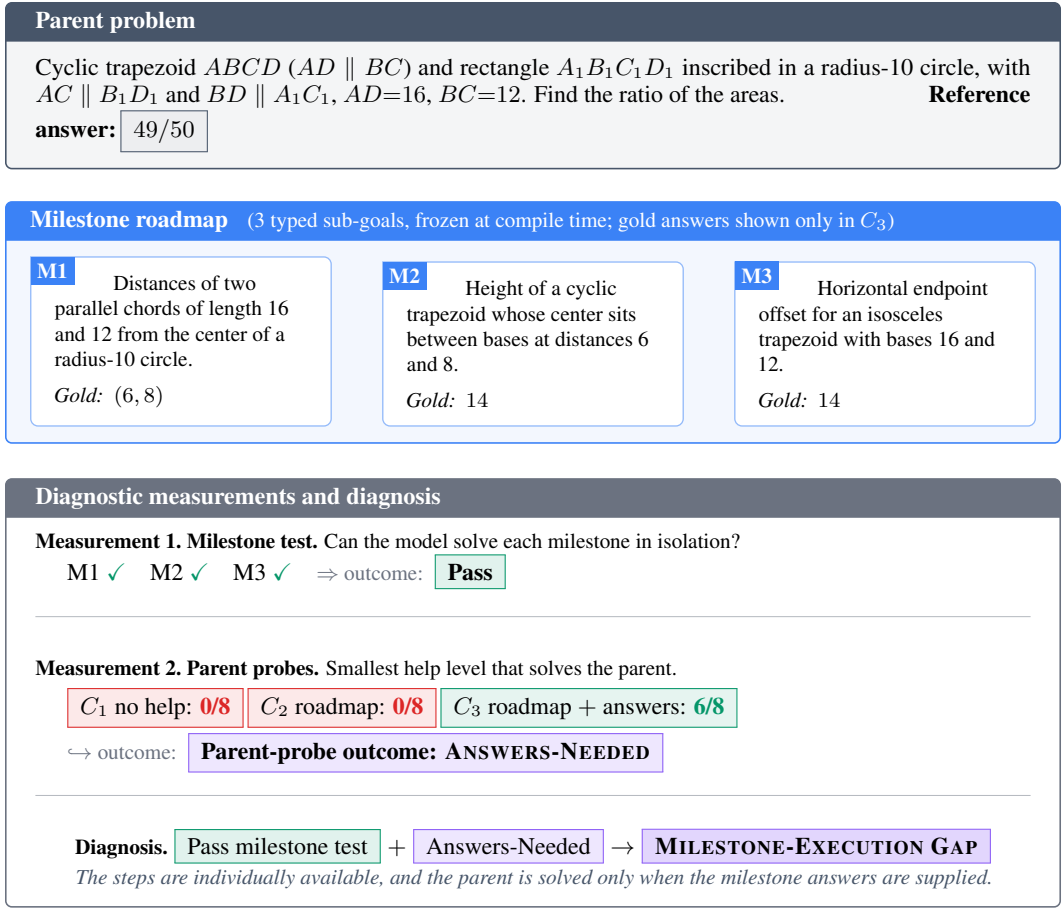

\centering
\begin{tcolorbox}[
  colback=wexParent!6, colframe=wexParent, boxrule=0.6pt, arc=2pt,
  left=8pt, right=8pt, top=3pt, bottom=3pt,
  title={\bfseries\small Parent problem},
  coltitle=white, colbacktitle=wexParent,
  fonttitle=\bfseries\small, width=\linewidth
]
\small Cyclic trapezoid $ABCD$ ($AD \parallel BC$) and rectangle $A_1 B_1 C_1 D_1$ inscribed in a radius-10 circle, with $AC \parallel B_1 D_1$ and $BD \parallel A_1 C_1$, $AD{=}16$, $BC{=}12$. Find the ratio of the areas. \hfill \textbf{Reference answer:}\,\fcolorbox{wexParent}{wexParent!10}{\,$49/50$\,}
\end{tcolorbox}

\vspace{1pt}

\begin{tcolorbox}[
  colback=wexMile!6, colframe=wexMile, boxrule=0.6pt, arc=2pt,
  left=6pt, right=6pt, top=3pt, bottom=3pt,
  title={\bfseries\small Milestone roadmap \;\;\textmd{\footnotesize (3 typed sub-goals, frozen at compile time; gold answers shown only in $C_3$)}},
  coltitle=white, colbacktitle=wexMile,
  fonttitle=\bfseries\small, width=\linewidth
]
\centering
\begin{minipage}[t]{0.30\linewidth}
\begin{tcolorbox}[colback=white, colframe=wexMile!55, boxrule=0.5pt, arc=2.5pt,
  left=6pt, right=6pt, top=4pt, bottom=4pt, enhanced, sharp corners=northwest, halign=flush left,
  overlay={\node[fill=wexMile, text=white, font=\bfseries\footnotesize, inner sep=2.5pt, anchor=north west] at (frame.north west) {M1};}]
\footnotesize\hspace{22pt}Distances of two parallel chords of length 16 and 12 from the center of a radius-10 circle.
\par\vspace{4pt}
\footnotesize\emph{Gold:}\, $(6,8)$
\end{tcolorbox}
\end{minipage}\hfill
\begin{minipage}[t]{0.30\linewidth}
\begin{tcolorbox}[colback=white, colframe=wexMile!55, boxrule=0.5pt, arc=2.5pt,
  left=6pt, right=6pt, top=4pt, bottom=4pt, enhanced, sharp corners=northwest, halign=flush left,
  overlay={\node[fill=wexMile, text=white, font=\bfseries\footnotesize, inner sep=2.5pt, anchor=north west] at (frame.north west) {M2};}]
\footnotesize\hspace{22pt}Height of a cyclic trapezoid whose center sits between bases at distances 6 and 8.
\par\vspace{4pt}
\footnotesize\emph{Gold:}\, $14$
\end{tcolorbox}
\end{minipage}\hfill
\begin{minipage}[t]{0.30\linewidth}
\begin{tcolorbox}[colback=white, colframe=wexMile!55, boxrule=0.5pt, arc=2.5pt,
  left=6pt, right=6pt, top=4pt, bottom=4pt, enhanced, sharp corners=northwest, halign=flush left,
  overlay={\node[fill=wexMile, text=white, font=\bfseries\footnotesize, inner sep=2.5pt, anchor=north west] at (frame.north west) {M3};}]
\footnotesize\hspace{22pt}Horizontal endpoint offset for an isosceles trapezoid with bases 16 and 12.
\par\vspace{4pt}
\footnotesize\emph{Gold:}\, $14$
\end{tcolorbox}
\end{minipage}
\end{tcolorbox}

\vspace{2pt}

\begin{tcolorbox}[
  colback=white, colframe=wexNeutral, boxrule=0.6pt, arc=2pt,
  left=8pt, right=8pt, top=3pt, bottom=4pt,
  title={\bfseries\small Diagnostic measurements and diagnosis},
  coltitle=white, colbacktitle=wexNeutral,
  fonttitle=\bfseries\small, width=\linewidth
]
{\footnotesize \textbf{Measurement 1.\ Milestone test.} Can the model solve each milestone in isolation?}
\par\vspace{2pt}
\hspace*{12pt}\small M1~\textcolor{wexCheck}{\checkmark}\quad M2~\textcolor{wexCheck}{\checkmark}\quad M3~\textcolor{wexCheck}{\checkmark}\quad{\footnotesize\color{wexNeutral}$\Rightarrow$\ outcome:}\, \fcolorbox{wexPass}{wexPass!12}{\,\textbf{Pass}\,}

\vspace{4pt}
{\color{wexNeutral!50}\rule{\linewidth}{0.3pt}}
\vspace{2pt}

{\footnotesize \textbf{Measurement 2.\ Parent probes.} Smallest help level that solves the parent.}
\par\vspace{2pt}
\hspace*{12pt}\fcolorbox{wexFail}{wexFail!10}{\,\small $C_1$ no help: \textcolor{wexFail}{\textbf{0/8}}\,}\,%
\fcolorbox{wexFail}{wexFail!10}{\,\small $C_2$ roadmap: \textcolor{wexFail}{\textbf{0/8}}\,}\,%
\fcolorbox{wexPass}{wexPass!10}{\,\small $C_3$ roadmap $+$ answers: \textcolor{wexPass}{\textbf{6/8}}\,}
\par\vspace{1pt}
\hspace*{12pt}{\footnotesize\color{wexNeutral}$\hookrightarrow$\ outcome:}\, \fcolorbox{wexAnswer}{wexAnswer!15}{\,\textbf{Parent-probe outcome: \textsc{Answers-Needed}}\,}

\vspace{4pt}
{\color{wexNeutral!50}\rule{\linewidth}{0.3pt}}
\vspace{2pt}

\centering
{\footnotesize\textbf{Diagnosis.}}\;\fcolorbox{wexPass}{wexPass!12}{\,Pass milestone test\,}\;$+$\;\fcolorbox{wexAnswer}{wexAnswer!15}{\,Answers-Needed\,}\;$\rightarrow$\;\fcolorbox{wexAnswer}{wexAnswer!25}{\,\textbf{\textsc{Milestone-Execution Gap}}\,}
\par\vspace{1pt}
{\footnotesize\itshape\color{wexParent} The steps are individually available, and the parent is solved only when the milestone answers are supplied.}
\end{tcolorbox}

\caption{\textbf{Worked example of the two diagnostic measurements} (Qwen-base on family \texttt{cabf0726}, numbers from the released 16K panel). The model solves all tested milestones in isolation, but the parent is solved only when milestone answers are supplied, yielding a \textsc{milestone-execution gap}.}
\label{fig:worked_example}
\end{figure}

OracleLadder fixes each parent problem and varies only the intermediate information shown to the evaluated model, which we call the student. For each student and milestone family, it measures two quantities: whether the student solves every tested milestone in isolation, and the smallest level of help that solves the parent problem. Figure~\ref{fig:worked_example} shows both measurements on one family.

\paragraph{Milestone compilation.}
Given a parent problem $P$, its reference solution $G(P)$, and a strong teacher model $T$, we compile a \emph{milestone family}: the parent problem together with a teacher-provided set of standalone milestones. Each milestone has a self-contained description, a milestone type, an answer schema, and a gold answer. The evaluated student is never used during compilation, so the milestone family depends only on $(P, G(P), T)$ and deterministic validation checks. The same family is held fixed across probes and reused unchanged across all evaluated models. Schema enforcement, anti-leakage filtering, milestone-type details, and family-level validity constraints are in Algorithm~\ref{alg:compiler} and Appendix~\ref{app:algorithms}.

\paragraph{Symbolic verification.}
We use deterministic symbolic verification to avoid LLM-judge grading bias. The verifier extracts the boxed answer, applies schema-aware parsing, and checks symbolic equivalence against the gold answer. The implementation uses symbolic algebra tools and \texttt{math-verify}~\citep{mathverify2025}. Appendix~\ref{app:verifier_ext} gives the parser cascade and a 400-response reliability audit, and Appendix~\ref{app:frontier_judge} compares the cascade with a frontier LLM judge.

\paragraph{Stage 0 milestone test.}
For each student and milestone family, we test whether the student can solve every tested milestone in isolation. The roadmap separates intermediate milestones, which develop the solution, from a final integration step whose gold answer is essentially the parent answer. The integration step is kept private, because showing or testing it would leak the parent answer. For each remaining milestone, we sample $K{=}8$ direct attempts and mark it as solved if at least one rollout passes symbolic verification. A family \emph{passes the milestone test} for a student if the student solves every tested milestone, and \emph{fails the milestone test} otherwise. Stage~0 is a per-student axis, separate from the three parent probes. The teacher decides how many milestones a roadmap has, and the milestone test becomes stricter as roadmaps become finer, because every milestone must be solved. We therefore report the milestone count together with milestone-test results, compare five-way labels only between roadmaps of matched granularity, and also report a variant scored by the fraction of milestones passed (Appendix~\ref{app:teacher_robust}).

\paragraph{Parent probing conditions and parent-probe outcome.}
For each milestone family, we evaluate the student under three levels of help on the parent problem: $C_1$ no help, $C_2$ parent plus milestone roadmap (intermediate milestones only, with the integration step kept private), $C_3$ parent plus roadmap plus gold milestone answers. Because the integration step is excluded, $C_3$ reveals the parent answer only when an intermediate milestone's gold answer is symbolically equivalent to it. A symbolic-equivalence check flags these cases, and Appendix~\ref{app:audit_details} reports their effect. The \emph{parent-probe outcome} is the smallest help level that produces at least one correct rollout in the ordered $C_1 \rightarrow C_2 \rightarrow C_3$ evaluation: \textsc{Direct}, \textsc{Roadmap-Needed}, \textsc{Answers-Needed}, or \textsc{Unrecovered}. Crossing this outcome with the milestone test yields the reasoning-gap diagnosis in Section~\ref{sec:lattice}.

\paragraph{Matched control conditions.}
We include three matched controls to test whether recovery comes from problem-matched milestone information rather than generic prompting or arbitrary answer content. $C_2$-random replaces the milestone roadmap with descriptions sampled from a different milestone family. $C_2$-generic replaces the specific roadmap with a generic decomposition prompt. $C_3$-mismatched keeps the correct roadmap but pairs it with wrong milestone answers sampled from other milestone families. Prompt templates and the probing procedure are in Algorithm~\ref{alg:diagnosis} and Appendix~\ref{app:prompts}, and Section~\ref{sec:oracle_specificity} reports the specificity results.
\section{Experimental Setup}
\label{sec:setup}

\paragraph{Diagnostic-set construction.}
We construct the diagnostic set from a 2{,}000-parent candidate pool sampled from NuminaMath-1.5-RL-Verifiable~\citep{numina2024}. The teacher model is GPT-5.4 Thinking~\citep{openai2026gpt54}. It successfully compiles valid milestone families for 1{,}024 of the 2{,}000 parents. We then use Qwen3-8B-Base~\citep{yang2025qwen3} before RL, which we call \emph{Qwen-base}, as the anchor model for the screening step. We use this checkpoint because the screen requires an intermediate-capability, pre-intervention model: too weak an anchor would fail isolated milestones, while too strong an anchor would solve most parents directly. Qwen-base is open-weight and is the common initialization for the two Qwen RL endpoints, so the diagnostic slice is fixed before either training intervention affects the screen. We screen the 1{,}024 valid families with Qwen-base at a 4K-token budget. A family passes the screen when Qwen-base fails the parent in all $K{=}8$ direct attempts but solves every tested non-\textsc{Integrate} milestone in at least one of $K{=}8$ attempts. This filter leaves 731 families, and requiring at least two tested non-\textsc{Integrate} milestones gives the $n{=}354$ main diagnostic set. The main panel re-samples every model at a 16K-token budget, where Qwen-base solves 76 of these families directly.

The set is therefore a stress set defined by the anchor and the source data. It targets families where the anchor shows a gap between the parent and its milestones, so category shares describe this set. A direct-failure screen with Qwen3-8B selects 7\% of MATH500 but 63\% of AIME 2024/25 (Appendix~\ref{app:second_datasets}), which shows how much the source data shapes such a set. Re-running the screen with each panel model as the anchor keeps the model ranking: all 15 pairwise Kendall comparisons of the induced rankings are positive, with mean $\tau{=}0.80$ under the 16K milestone test and $0.62$ under the 4K test (Appendix~\ref{app:dataset}). Appendix~\ref{app:dataset} gives the funnel counts, and Appendix~\ref{app:heldout} a 32-family held-out slice compiled from a separate 150-parent NuminaMath sample.

\paragraph{Model panel.}
We evaluate six model instances from four model lineages. Three are from the Qwen3-8B lineage: Qwen-base and two RL endpoints initialized from it, OutcomeRL-2K and MilestoneRL-2K. The suffix ``2K'' denotes the 2{,}000-parent RL training pool. The two endpoints share the same base checkpoint, LoRA rank 64~\citep{hu2022lora}, GRPO recipe~\citep{shao2024deepseekmath}, AdamW optimizer settings, decoding budget, and 180-step training horizon. They are trained on a separate 2{,}000-parent NuminaMath training pool, which shares no parent problems with the 354 diagnostic families, as verified by parent identifiers in the released artifact.

The endpoints differ in reward construction and auxiliary training details. OutcomeRL-2K rewards parent-level correctness only. MilestoneRL-2K uses the same parent-level reward and additionally trains with milestone-level sub-goal rollouts whose correctness contributes to the reward. The milestone run also differs in per-step rollout budget, SFT warmup, and curriculum (Appendix~\ref{app:training_configs}), and Section~\ref{sec:results_training} explains how we read the comparison. The remaining three lineages contribute one model each: gpt-oss-20b~\citep{openai2025gptoss}, Llama-3.3-70B-Instruct~\citep{meta2024llama33}, and DeepSeek-V3.1~\citep{deepseek2024v3}. Full model and serving metadata are in Appendix~\ref{app:panel}.

\paragraph{Evaluation protocol and aggregate comparator.}
Unless otherwise stated, all parent-probe experiments use the main evaluation setting: $K{=}8$ rollouts per condition, \texttt{max\_tokens=16384}, identical prompts, and the same symbolic verifier. A family is marked solved under a condition if at least one of the $K$ rollouts verifies as correct. Stage~0 screening during construction used \texttt{max\_tokens=4096}, and all main-text taxonomy and panel results use the 16K Stage~0 rerun. Appendix~\ref{app:stage0} compares the two budgets. We also report MATH hard-100 as an external scalar comparator: a 100-problem subset of level-5 MATH~\citep{hendrycksmath2021}, evaluated under the same decoding budget and reported as mean single-rollout correctness over $K{=}8$ samples. Details are in Appendix~\ref{app:aggregate_settings}.

\begin{table}[!b]
\centering
\footnotesize
\caption{\textbf{Same-set summary on the 354 milestone families.}
The table separates direct solving, diagnostic recovery, and external MATH performance.}
\label{tab:same_set_summary}
\vspace{3pt}
\setlength{\tabcolsep}{4pt}
\begin{tabular}{@{}lccccc@{}}
\toprule
\textbf{Model} &
\shortstack{\textbf{Direct}\\\textbf{accuracy}} &
\shortstack{\textbf{Solved}\\\textbf{directly}} &
\shortstack{\textbf{Solved by}\\\textbf{any probe}} &
\shortstack{\textbf{Added by}\\\textbf{milestones}} &
\shortstack{\textbf{MATH}\\\textbf{score}} \\
\midrule
Qwen-base       &  5.4\% & 21.5\% & 42.9\% & 21.5\% & 45.4\% \\
OutcomeRL-2K    & 10.7\% & 29.9\% & 48.0\% & 18.1\% & 65.4\% \\
MilestoneRL-2K  & 10.6\% & 27.1\% & 46.6\% & 19.5\% & 63.4\% \\
gpt-oss-20b     & 25.7\% & 45.5\% & 60.5\% & 15.0\% & 88.6\% \\
Llama-70B       &  7.8\% & 14.7\% & 35.6\% & 20.9\% & 49.5\% \\
DeepSeek-V3.1   & 28.4\% & 40.4\% & 51.7\% & 11.3\% & 85.5\% \\
\bottomrule
\end{tabular}

\vspace{3pt}
\parbox{0.96\linewidth}{\footnotesize
\emph{Notes.} Direct accuracy is per-rollout parent-only success. The other diagnostic columns are family-level shares on the 354 milestone families with a $\geq 1/8$ threshold on $K{=}8$ rollouts. Solved by any probe includes direct, roadmap, and roadmap-plus-answers solves. Added by milestones is the gain over solved directly. MATH score is the external hard-100 comparator.
}
\end{table}

\section{Results}
\label{sec:results}

\paragraph{Outline.}
We first check that recovery comes from problem-matched milestone information, on the main set and on new datasets. We then use OracleLadder across the model panel and across training, where two RL runs start from Qwen-base. The central finding is that \textsc{composition-gap} is the largest gap for every model. We close by measuring what this category contains.

\subsection{Recovery Comes from Problem-Matched Milestones}
\label{sec:oracle_specificity}

\begin{figure}[!ht]
\centering
\includegraphics[width=\textwidth]{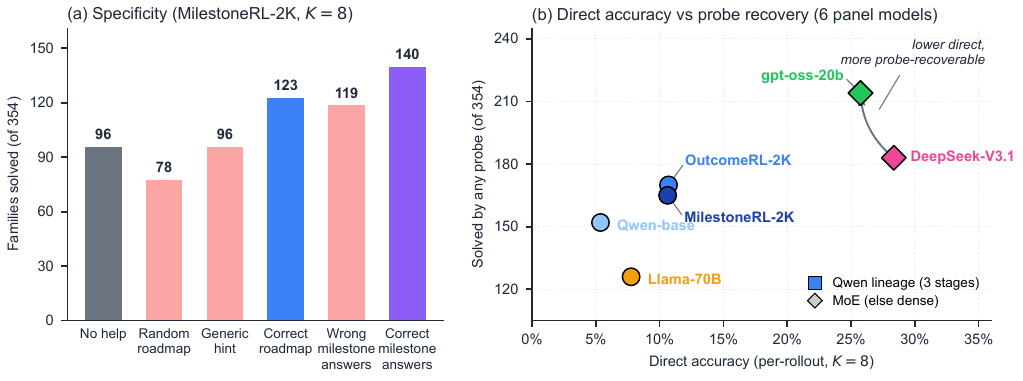}
\caption{\textbf{Problem-matched milestones drive recovery, and scalar accuracy does not determine diagnostic recovery.} (a)~MilestoneRL-2K specificity check: correct milestone information vs.\ random, generic, and mismatched controls. (b)~Cross-model: per-rollout direct accuracy vs.\ family-level any-probe recovery on the same 354 families.}
\label{fig:specificity_decoupling}
\end{figure}

Figure~\ref{fig:specificity_decoupling}(a) checks whether recovery is caused by the specific milestone information or merely by giving the model a longer decomposition-style prompt. For MilestoneRL-2K, the correct-roadmap condition ($C_2$-correct) and the roadmap-plus-answers condition ($C_3$-gold) solve substantially more families than direct prompting, while random roadmaps and generic decomposition prompts do not improve over the direct baseline. The mismatched-answer control keeps the correct roadmap but corrupts the milestone answers. Its lift over direct prompting suggests that the roadmap itself contributes useful structure, while its gap to $C_3$-gold shows that problem-matched milestone answers add further information.

\begin{table}[t]
\centering
\small
\caption{\textbf{The roadmap effect replicates beyond the main set.} Families solved under each condition. The main set uses MilestoneRL-2K. The new datasets use Qwen3-8B (thinking disabled) as screening anchor and student. AIME uses the default threshold of at least 1 correct attempt out of 8. MATH500 uses at least 3 of 8, because at the default threshold direct prompting already recovers 16 of its 35 families, which passed a screen of only 4 attempts (Appendix~\ref{app:second_datasets} gives every threshold).}
\label{tab:replication}
\vspace{3pt}
\begin{tabular}{@{}lccc@{}}
\toprule
\textbf{Condition} & \textbf{Main set} ($n{=}354$) & \textbf{MATH500} ($n{=}35$) & \textbf{AIME 2024/25} ($n{=}38$) \\
\midrule
$C_1$ direct & 96 & 3 & 2 \\
$C_2$-random & 78 & 2 & 5 \\
$C_2$-generic & 96 & 2 & 5 \\
\textbf{$C_2$-correct} & \textbf{123} & \textbf{15} & \textbf{17} \\
$C_3$-mismatched & 119 & 13 & 14 \\
\textbf{$C_3$-gold} & \textbf{140} & \textbf{15} & \textbf{15} \\
\bottomrule
\end{tabular}
\end{table}

\paragraph{The contrast replicates on new datasets.} We ran the full protocol on MATH500 and AIME 2024/25 with the same teacher model, parent-probe prompts, verifier, and $K$. The screen, the student, and the teacher's input differ from the 354-family set, and Appendix~\ref{app:second_datasets} lists each difference. On both datasets the matched roadmap solves far more families than either corrupted roadmap (Table~\ref{tab:replication}). A 10-problem code-generation pilot on LiveCodeBench, graded by running the tests, shows the same ladder of help: 6.5\%, 41.2\%, and 88.8\% of attempts pass with no help, with the roadmap, and with the roadmap plus verified helper functions (Appendix~\ref{app:code_pilot}). A held-out NuminaMath slice of 32 families shows the specificity pattern as well (Appendix~\ref{app:heldout}).

\subsection{Cross-Model Readout: Direct Accuracy and Diagnostic Recovery Decouple}
\label{sec:cross_model}

The first use of the diagnostic is cross-model: compare models on the same milestone families by failure structure, not only by scalar accuracy. Table~\ref{tab:same_set_summary} and Figure~\ref{fig:specificity_decoupling}(b) compare per-rollout direct accuracy with the family-level count of problems solved by any diagnostic probe. The two strongest models trade places depending on the measure. DeepSeek-V3.1 has the highest per-rollout direct accuracy (28.4\%), yet gpt-oss-20b solves more families directly (45.5\% vs.\ 40.4\%) and more families by any probe (60.5\% vs.\ 51.7\%). Milestone help adds 15.0\% of families for gpt-oss-20b and 11.3\% for DeepSeek-V3.1, so gpt-oss-20b leaves more of its direct failures recoverable. A single accuracy score hides this difference. In family-level bootstrap resampling, gpt-oss-20b solves the most families by any probe in all 1{,}000 resamples (Appendix~\ref{app:bootstrap}).

\subsection{Reasoning-Gap Taxonomy: Composition Is the Largest Gap}
\label{sec:lattice}

\begin{figure}[t]
\centering
\includegraphics[width=\textwidth]{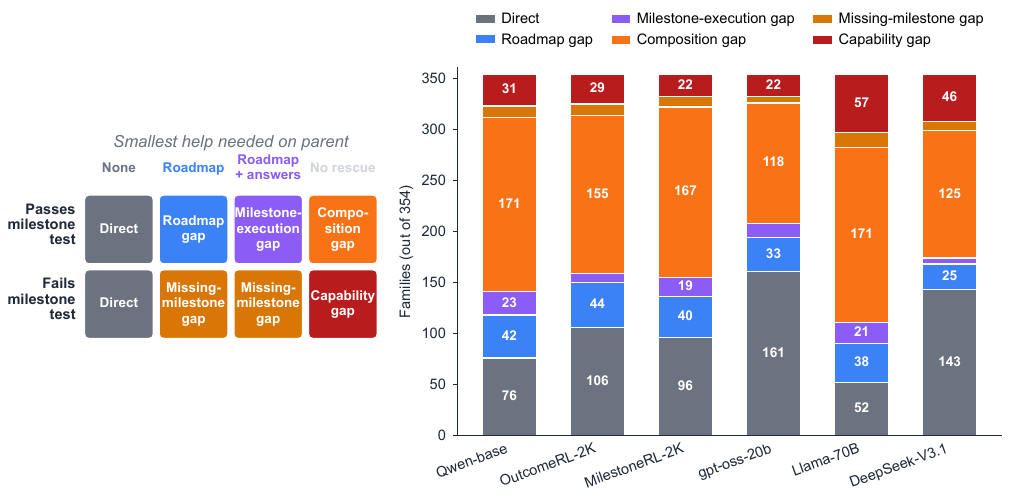}
\caption{\textbf{Largest non-direct category across all six models: pass the milestone test, but no parent probe yields a correct rollout.}
Reasoning-gap taxonomy on the 354-family diagnostic set. See Section~\ref{sec:repair} for the audit reading.}
\label{fig:lattice}
\end{figure}

The parent-probe outcome says how much help was needed. The reasoning-gap taxonomy asks where the failure lies. We cross the milestone test with the smallest help level that solves the parent. Families that pass the milestone test split into three: \textsc{roadmap-gap} when the roadmap solves the parent, \textsc{milestone-execution-gap} when the roadmap-plus-answers condition is needed, and \textsc{composition-gap} when no parent probe solves the parent. Families that fail the milestone test split into two: \textsc{missing-milestone-gap} when at least one of $C_2$ or $C_3$ recovers the parent (so the missing milestone-level step is supplied through the roadmap or its answers) and \textsc{capability-gap} when no parent probe does. Figure~\ref{fig:lattice} shows the cross, and Appendix~\ref{app:stage0} gives per-model counts and budget sensitivity.

\paragraph{Composition is the largest gap.}
\textsc{Composition-gap} is the largest of the five gaps for every model, covering 33--48\% of families (Figure~\ref{fig:lattice}). The category is defined by what the two tests measure: the student solves every tested milestone alone, and no parent probe solves the parent. Section~\ref{sec:repair} measures what it contains.

\paragraph{The taxonomy localizes cross-model differences.}
The other gap categories vary in ways that the four-state parent-probe outcomes alone do not expose. \textsc{Capability-gap} varies sharply, from 6--9\% for the Qwen lineage and gpt-oss-20b to 13\% for DeepSeek-V3.1 and 16\% for Llama-3.3-70B. \textsc{Roadmap-gap} is more uniform at 7--12\%, while \textsc{milestone-execution-gap} is small for every model at 2--7\%. Per-model taxonomy counts are in Table~\ref{tab:bottleneck_lattice}, Appendix~\ref{app:stage0}.

\paragraph{Robustness of the readout.}
Per-family recovery changes little when the roadmaps change. When an independent teacher (Inkling) writes the roadmaps for 53 families, gpt-oss-20b gets the same $C_2$ outcome on 83\%, the same $C_3$ outcome on 87\%, and the same joint outcome on 77\% of them. When the original teacher writes roadmaps 4.1 times finer, agreement on 33 families is 85\%, 85\%, and 76\%. Five-way labels are more sensitive, because the milestone test requires every milestone of the roadmap, and the second teacher's roadmaps have more milestones that the student solves less often alone. Across teachers they agree on 29 of 39 families when the milestone test is scored by the fraction of milestones passed, and on 22 of 39 under the strict test (Appendix~\ref{app:teacher_robust}). Within the 354 set, re-screening with each panel model as the anchor keeps the model ranking (mean Kendall $\tau{=}0.80$, Appendix~\ref{app:dataset}). The state ordering is the same at every success threshold (Appendix~\ref{app:threshold}), and \textsc{composition-gap} counts move by at most 7 families between the 4K and 16K Stage~0 budgets (Appendix~\ref{app:stage0}). The roadmaps also match what models write on their own. Milestone answers appear in the models' unaided solutions at 2.9 to 3.9 times the rate of a cross-problem control, and at 6.6 to 12.6 times for answers that the problem statement does not already contain (Appendix~\ref{app:trace}).

\subsection{Longitudinal Readout: Similar RL Gains, Different State Movement}
\label{sec:results_training}

We now use the same readout longitudinally and ask what RLVR changes when two runs start from the same base model.

Both runs start from Qwen-base and train on the shared pool described in Section~\ref{sec:setup}, which is disjoint from the 354 diagnostic families. OutcomeRL-2K rewards parent correctness only. MilestoneRL-2K also rewards milestone-level rollouts, starts from an SFT warmup on R1-distilled milestone solutions, and uses half the per-step rollout budget (Table~\ref{tab:training_configs}). The runs differ in several ways and are not matched in compute, so we read them as two readouts of training, and the comparison does not rank the training methods.

\begin{figure}[t]
\centering
\includegraphics[width=\textwidth]{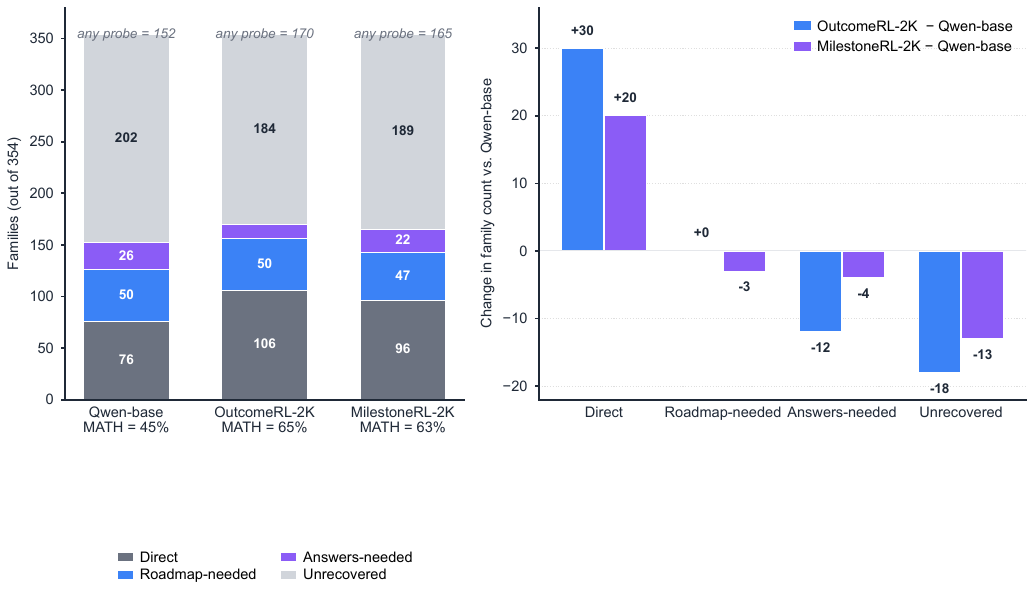}
\caption{\textbf{Similar RL gains can move families through different parent-probe outcomes.}
(Left)~Parent-probe outcomes for Qwen-base and the two RL runs. (Right)~Change in each outcome relative to Qwen-base.}
\label{fig:two_arm_redistribution}
\end{figure}

Figure~\ref{fig:two_arm_redistribution} shows that the two runs reach similar MATH hard-100 gains (+20.0 and +18.0 points) but change the parent-probe outcomes differently. Relative to Qwen-base, OutcomeRL-2K has 30 more \textsc{Direct} and 12 fewer \textsc{Answers-Needed} families, while MilestoneRL-2K has 20 more and 4 fewer. A single accuracy score on either set hides this difference. Appendices~\ref{app:transition},~\ref{app:trajectory}, and~\ref{app:decoding_budget} give the family-level transitions between the two runs, intermediate checkpoints, and decoding-budget sensitivity.

\subsection{What the Composition Gap Contains}
\label{sec:repair}

We report three quantities separately. The first is the \textsc{composition-gap} share itself, 33--48\% of families per model (Section~\ref{sec:lattice}).

The second is the share of this category caused by grading. An LLM rubric review (GPT-5.5) of every model's \textsc{composition-gap} families flags 22--29\% per model as verifier or format errors, for example a correct answer the cascade cannot parse or a defective reference answer. Removing them leaves 24--37\% of families in the category for every model, and only models within one percentage point of each other change places (Appendix~\ref{app:audit_rebuttal}). A frontier judge points the same way. On 997 paired responses, gpt-5.5 accepts 12.7\% that the cascade rejects and rejects 0.1\% that the cascade accepts. A grader-blind adjudication of 34 disagreements traces about two thirds of them to cascade strictness or defective reference answers and one third to judge over-acceptance, mostly credit for unfinished work (Appendix~\ref{app:frontier_judge}).

The third is direct evidence of composition failure. We re-ran the 18 families that the author audit labels as genuine composition failures (Appendix~\ref{app:audit_details}) at $K{=}8$ with the roadmap and every gold milestone answer, saved the raw outputs, and graded them by hand. With every milestone answer given, Qwen3-8B (thinking disabled), a different 8B model from the audited student, solves 16 of the 18 at least once. On one family it solves both milestones alone in 8 of 8 attempts, receives both answers, and still fails all eight attempts, while gpt-oss-20b solves it from the same roadmap and answers in 6 of 8. The retest also finds defective reference answers in 5 of the 18 families and wrong gold milestone answers in 1, which is further evidence for the grading layer above (Appendix~\ref{app:audit_rebuttal}). The author audit of 100 unrecovered MilestoneRL-2K families also finds roadmaps that omit a bridging step (Appendix~\ref{app:audit_details}). The category therefore mixes grading errors, which we now measure for every model, roadmaps that miss a step, and genuine failures to combine steps the model can already do, which the raw outputs show directly.
\section{Related Work}
\label{sec:related}

\paragraph{Math benchmarks.}
Grade-school benchmarks include GSM8K, SVAMP, and ASDiv~\citep{Cobbe2021VerifiersGSM8K,patel2021svamp,miao2020asdiv}. GSM-Hard, GSM-Plus, and GSM-Symbolic perturb or harden them~\citep{gao2023pal,li2024gsmplus,mirzadeh2024gsmsymbolic}. Competition suites include MATH, NuminaMath, and TheoremQA~\citep{hendrycksmath2021,numina2024,chen2023theoremqa}, and SciBench and JEEBench add science and engineering exams~\citep{wang2024scibench,arora2023jeebench}. Olympiad sets add harder problems~\citep{he2024olympiadbench,gao2024omnimath,huang2024olympicarena}, and FrontierMath and PutnamBench reach research level~\citep{glazer2024frontiermath,tsoukalas2024putnambench}. Scores on the lower tiers are saturating for frontier systems~\citep{deepseek2025r1}, and an aggregate score compresses different failures into one number~\citep{Dehghani2021BenchmarkLottery}. OracleLadder adds a diagnostic layer over existing problems, so the same data also shows which reasoning step remains the bottleneck.

\paragraph{Structured and intervention-based evaluation.}
Behavioral test suites replace a single score with structured tests~\citep{ribeiro2020checklist,srivastava2023bigbench,suzgun2023bbh}. Holistic benchmarks cover many tasks at once~\citep{liang2023helm,hendrycks2021mmlu}, and documentation standards state what a dataset or model is for~\citep{mitchell2019modelcards,gebru2021datasheets}. LLM-judged step quality~\citep{xia2024reasoneval} captures more than the final answer but inherits judge bias~\citep{Zheng2023MTBench}, so we grade with a deterministic verifier. Other studies change what the model sees. Problem-level counterfactuals~\citep{wu2024reasoning,shi2023distractors} alter the problem, and trace-level edits~\citep{lanham2023measuring,turpin2023language} alter the reasoning the model conditions on. Prompting that writes out steps or sub-problems improves accuracy~\citep{Wei2022CoT,Kojima2022ZeroShotReasoners,Zhou2022LeastToMost}. CHAMP~\citep{mao2024champ} and \citet{agrawal2024hint} measure how much problem-specific hints help on math, and \citet{qiu2024arranging} identify arranging the steps as a bottleneck of chain-of-thought reasoning. We-Math~\citep{qiao2025wemath} tests whether multimodal models that solve the sub-problems of a visual math problem also solve the whole. \citet{press2023compositionality} name the compositionality gap in multi-hop question answering, and \citet{hosseini2024notall} find a related gap when two grade-school problems are chained. OracleLadder combines a fixed roadmap, graded levels of help, and a separate test of each milestone, which separates a missing step from steps that are available but not combined. Decomposition-search methods consider many decompositions per problem~\citep{khot2022decomposed,yao2023tot}. We hold one teacher roadmap fixed per problem, so any lift reflects problem-matched content.

\paragraph{Process supervision and RL training.}
Step-level scorers grade or train chain-of-thought traces~\citep{uesato2022solving,lightman2023verify,wang2024mathshepherd}. Later work improves how they are trained~\citep{luo2024omegaprm,she2025rprm,zhang2025lessons}, and benchmarks test them directly~\citep{zheng2024processbench,song2025prmbench}. Test-time-compute methods use such scorers as guides at inference~\citep{snell2024scaling,brown2024monkeys}. The taxonomy suggests where step-level feedback could help most: \textsc{roadmap-gap} and \textsc{milestone-execution-gap} families, where the bottleneck is planning or executing a step. In \textsc{composition-gap} families the model already has the milestone answers, so the remaining work is combining them. RL methods for math report aggregate gains~\citep{shao2024deepseekmath,yu2025dapo,deepseek2025r1}, and so do related self-improvement methods~\citep{guan2025rstarmath,hou2025treerl,ma2025s2r,yuan2023scaling}. \citet{qin2026turnover} track which problems RLVR training solves and later loses. The longitudinal readout adds which gap each problem moves through.

\section{Discussion and Limitations}
\label{sec:discussion}

OracleLadder is a diagnostic instrument for comparing models and training runs on a fixed set of problems. It supports two uses: comparing models by where they fail on the same families, and tracking how training moves families between states. Counts such as families solved by any probe describe this stress set, so they support comparisons between models and conditions on it.

\paragraph{What the readout implies.}
Across all six models, the largest gap is one where the model solves every milestone alone yet no level of help solves the problem. The category mixes grading errors, roadmaps that omit a bridging step, and genuine failures to combine available steps (Section~\ref{sec:repair}). Measuring these parts separately tells a developer where to act: the verifier and reference answers for the first, roadmap coverage for the second, and training that targets the final combination of already-correct steps for the third.

\paragraph{Limitations.}
OracleLadder needs an automatic verifier for final and intermediate answers. We use a symbolic cascade for math and test execution in the code pilot. The 354-family set is screened by Qwen-base, and category shares depend on the anchor, the source dataset, and the student, while the matched-versus-corrupted contrast replicates across datasets (Appendix~\ref{app:second_datasets}). Each family has one teacher roadmap. A second teacher and a fourfold change in granularity leave per-family recovery largely unchanged, and five-way labels need roadmaps of matched granularity (Appendix~\ref{app:teacher_robust}). Two authors ran the audit without external blinding, and its raw outputs are released for inspection. NuminaMath and MATH are public, so absolute scores carry the usual contamination caveat. The design compares conditions on the same problems, which limits how much contamination can move the contrasts. Compiling new roadmaps needs a strong teacher model, while any new student can be evaluated on the released roadmaps directly.

\paragraph{Release.}
The release contains the 354-family diagnostic set with roadmaps, the six-model Stage~0 and parent-probe panels at 16K, the verifier, all prompts, audit labels, leak-safe repair logs, and the replication data for MATH500, AIME 2024/25, and the code pilot. Some appendix analyses (the 4K-vs-16K audit, the format audit, intermediate checkpoints, and the held-out slice) depend on intermediate logs, and the release ships their summary tables. The MATH hard-100 comparator is released as per-model outputs keyed by public MATH identifiers. Code is released under the MIT license and data under CC BY 4.0.

\begin{ack}
Most of the training and inference in this work was funded by a Tinker research grant from Thinking Machines Lab, with additional compute credits from Prime Intellect. Both were provided through a course at the Massachusetts Institute of Technology (MIT). Thinking Machines Lab also develops Inkling, which we use as a second teacher in Appendix~\ref{app:teacher_robust}. The authors declare no other competing interests.
\end{ack}

\newpage
\bibliography{refs}
\bibliographystyle{plainnat}

\newpage
\appendix

\section{Algorithms}
\label{app:algorithms}

This appendix gives pseudo-code for the three implementation steps behind the diagnostic workflow: milestone-family compilation, the Stage~0 milestone test, and parent-probe outcome assignment. The algorithms are reproducibility specifications. The methodological contribution is the diagnostic design described in the main text.

\begin{algorithm}[h]
\caption{Milestone Family Compilation (student-free)}
\label{alg:compiler}
\begin{algorithmic}[1]
\REQUIRE Parent problem $P$, source reference solution $G(P)$, teacher model $\mathcal{T}$
\ENSURE Valid milestone family $\mathcal{F}(P)$ of typed, symbolically-checkable milestones
\REPEAT
    \STATE $\mathcal{F} \gets \mathcal{T}.\text{decompose}(P, G(P))$ \hfill $\triangleright$ \textit{propose candidate family}
    \FOR{each milestone $M_i \in \mathcal{F}$}
        \STATE \textit{schema}$_i \gets \mathcal{T}.\text{assign\_schema}(M_i)$
        \STATE $a_i \gets \mathcal{T}.\text{solve}(M_i)$
        \IF{$\neg\;\text{SchemaCheck}(a_i, \textit{schema}_i)$}
            \STATE mark $M_i$ invalid
        \ENDIF
    \ENDFOR
    \STATE filter milestones whose answer restates the parent solution (best-effort leak check, residual reported in Appendix~\ref{app:audit_details})
    \STATE validate family-level constraints: standalone solvability, non-degeneracy, $\geq 2$ milestones
\UNTIL{$\mathcal{F}$ satisfies all validity checks}
\RETURN $\mathcal{F}(P) = \{(M_i, \textit{schema}_i, a_i)\}$
\end{algorithmic}
\end{algorithm}

\begin{algorithm}[h]
\caption{Stage 0 Milestone Test}
\label{alg:stage0}
\begin{algorithmic}[1]
\REQUIRE Family $\mathcal{F}(P) = \{M_1, \ldots, M_k\}$, student $\mathcal{S}$, rollouts $K{=}8$
\ENSURE Per-milestone solve rate $\hat p(M_i)$ and milestone-test-pass flag $v_\mathcal{S}$
\FOR{each $M_i \in \mathcal{F}$ with \textit{type}$_i \neq \textsc{Integrate}$}
    \STATE $\hat{p}(M_i) \gets \frac{1}{K} \sum_{j=1}^{K} \mathbb{1}[\text{Verify}(\mathcal{S}(M_i)_j, a_i)]$
\ENDFOR
\STATE $v_\mathcal{S} \gets \bigwedge_{i: \textit{type}_i \neq \textsc{Integrate}} \mathbb{1}[\hat{p}(M_i) \geq 1/K]$
\RETURN $\{\hat{p}(M_i)\}, v_\mathcal{S}$
\end{algorithmic}
\end{algorithm}

\begin{algorithm}[h]
\caption{Parent-Probe Outcome Assignment}
\label{alg:diagnosis}
\begin{algorithmic}[1]
\REQUIRE Family $\mathcal{F}(P)$, student $\mathcal{S}$, $K{=}8$
\ENSURE State $s \in \{\textsc{Direct}, \textsc{Roadmap-Needed}, \textsc{Answers-Needed}, \textsc{Unrecovered}\}$
\STATE $n_1 \gets |\{j : \text{Verify}(\mathcal{S}(P)_j, a_P)\}|$ \hfill \textit{(C1)}
\STATE $n_2 \gets |\{j : \text{Verify}(\mathcal{S}(P \oplus \text{desc})_j, a_P)\}|$ \hfill \textit{(C2)}
\STATE $n_3 \gets |\{j : \text{Verify}(\mathcal{S}(P \oplus \text{desc} \oplus \text{gold})_j, a_P)\}|$ \hfill \textit{(C3)}
\IF{$n_1 \geq 1$} \RETURN \textsc{Direct}
\ELSIF{$n_2 \geq 1$} \RETURN \textsc{Roadmap-Needed}
\ELSIF{$n_3 \geq 1$} \RETURN \textsc{Answers-Needed}
\ELSE \RETURN \textsc{Unrecovered}
\ENDIF
\end{algorithmic}
\end{algorithm}

\section{Dataset Funnel and Milestone Types}
\label{app:dataset}
\label{app:milestone_types_table}

This appendix details the construction filters that turn the 2{,}000-parent NuminaMath sample into the $n{=}354$ diagnostic set, and reports the compiled milestone-type distribution. Three filters are applied in series: teacher compilation (1{,}024/2{,}000), anchor-solvable plus Stage~0 milestone-test pass under Qwen-base ($731/1{,}024$), and $\geq 2$ non-\textsc{Integrate} milestones ($354/731$). The set is therefore anchor-conditioned on Qwen-base screening, by design, and the released artifact lets users rerun the screen with a different anchor model.

\paragraph{Re-anchoring the screen.} We run the screen with each of the six panel models as the anchor in turn. For each anchor, we keep the families in the 354 set where that anchor passes Stage~0 and fails the parent under direct prompting. For each anchor's subset, we rank the five other models by their \textsc{composition-gap} count and compute Kendall's $\tau$ between every pair of anchor-induced rankings. With the 16K milestone test used throughout the paper, subsets range from 156 to 236 families and all 15 anchor pairs have $\tau > 0$, with mean $+0.80$ (Tables~\ref{tab:anchor_compgap_counts_16k} and~\ref{tab:anchor_kendall_16k}). Under each of the other four anchors, gpt-oss-20b and DeepSeek-V3.1 have the two smallest counts. The 4K milestone test used during construction gives mean $+0.62$, also with all 15 pairs positive (Tables~\ref{tab:anchor_compgap_counts} and~\ref{tab:anchor_kendall}).

\begin{table}[h]
\centering
\footnotesize
\caption{Per-anchor \textsc{composition-gap} counts with the 16K milestone test. Rows are anchor models, with the anchored subset size in parentheses. The anchored subset holds the families of the 354-family set where the anchor passes the milestone test and fails all 8 $C_1$ attempts. Each cell counts the families in that subset where the column model passes the milestone test and is \textsc{Unrecovered} under all three parent probes. A dash marks the anchor's own column, which is left unscored. The milestone test and the parent probes both run at \texttt{max\_tokens=16384} with $K{=}8$.}
\label{tab:anchor_compgap_counts_16k}
\vspace{4pt}
\setlength{\tabcolsep}{4pt}
\begin{tabular}{lcccccc}
\toprule
\textbf{Anchor (subset)} & \shortstack{\textbf{Qwen-}\\\textbf{base}} & \shortstack{\textbf{Outcome-}\\\textbf{RL-2K}} & \shortstack{\textbf{Milestone-}\\\textbf{RL-2K}} & \shortstack{\textbf{gpt-oss-}\\\textbf{20b}} & \shortstack{\textbf{Llama-}\\\textbf{70B}} & \shortstack{\textbf{DeepSeek-}\\\textbf{V3.1}} \\
\midrule
Qwen-base ($n{=}236$) & --- & 140 & 152 & 102 & 142 & 99 \\
OutcomeRL-2K ($n{=}208$) & 144 & --- & 150 & 100 & 132 & 101 \\
MilestoneRL-2K ($n{=}226$) & 152 & 145 & --- & 103 & 139 & 104 \\
gpt-oss-20b ($n{=}165$) & 108 & 103 & 114 & --- & 97 & 96 \\
Llama-70B ($n{=}230$) & 135 & 123 & 126 & 89 & --- & 91 \\
DeepSeek-V3.1 ($n{=}156$) & 103 & 100 & 106 & 86 & 96 & --- \\
\bottomrule
\end{tabular}
\end{table}

\begin{table}[h]
\centering
\footnotesize
\caption{Kendall's $\tau$ between prober rankings under two anchor screens, with the 16K milestone test. Within each anchor screen, the five other models are ranked by their \textsc{composition-gap} counts in Table~\ref{tab:anchor_compgap_counts_16k}. Each cell compares two anchors on the four models that are anchor in neither screen. The diagonal is 1 by definition. The mean over the 15 anchor pairs is $+0.80$, and all 15 are positive.}
\label{tab:anchor_kendall_16k}
\vspace{4pt}
\setlength{\tabcolsep}{4pt}
\begin{tabular}{lcccccc}
\toprule
\textbf{Anchor} & \shortstack{\textbf{Qwen-}\\\textbf{base}} & \shortstack{\textbf{Outcome-}\\\textbf{RL-2K}} & \shortstack{\textbf{Milestone-}\\\textbf{RL-2K}} & \shortstack{\textbf{gpt-oss-}\\\textbf{20b}} & \shortstack{\textbf{Llama-}\\\textbf{70B}} & \shortstack{\textbf{DeepSeek-}\\\textbf{V3.1}} \\
\midrule
Qwen-base & 1.00 & +0.67 & +0.33 & +0.67 & +0.67 & +0.67 \\
OutcomeRL-2K & +0.67 & 1.00 & +1.00 & +1.00 & +0.67 & +1.00 \\
MilestoneRL-2K & +0.33 & +1.00 & 1.00 & +1.00 & +1.00 & +1.00 \\
gpt-oss-20b & +0.67 & +1.00 & +1.00 & 1.00 & +0.67 & +1.00 \\
Llama-70B & +0.67 & +0.67 & +1.00 & +0.67 & 1.00 & +0.67 \\
DeepSeek-V3.1 & +0.67 & +1.00 & +1.00 & +1.00 & +0.67 & 1.00 \\
\bottomrule
\end{tabular}
\end{table}

\begin{table}[h]
\centering
\footnotesize
\caption{Per-anchor composition-gap counts. Rows are anchor models, and columns are non-anchor probers. Anchored subset size in parentheses next to the anchor name. The cell is the count of families in the anchored subset for which the prober passes Stage 0 and is \textsc{Unrecovered} under all three parent probes. Anchor and prober Stage 0 readings are both at the 4K screening budget (the budget used during diagnostic-set construction) to keep anchor-vs-prober comparable.}
\label{tab:anchor_compgap_counts}
\vspace{4pt}
\begin{tabular}{lcccccc}
\toprule
\textbf{Anchor (subset)} & \textbf{Qwen-b.} & \textbf{Outc-2K} & \textbf{Mile-2K} & \textbf{gpt-oss} & \textbf{Llama-70B} & \textbf{DSeek} \\
\midrule
Qwen-b. ($n{=}235$) & --- & 143 & 150 & 103 & 144 & 104 \\
Outc-2K ($n{=}209$) & 147 & --- & 146 & 102 & 126 & 101 \\
Mile-2K ($n{=}223$) & 154 & 144 & --- & 106 & 135 & 104 \\
gpt-oss ($n{=}163$) & 110 & 103 & 115 & --- & 91 & 99 \\
Llama-70B ($n{=}222$) & 137 & 119 & 123 & 84 & --- & 86 \\
DSeek ($n{=}154$) & 103 & 97 & 104 & 89 & 88 & --- \\
\bottomrule
\end{tabular}
\end{table}

\begin{table}[h]
\centering
\footnotesize
\caption{Kendall's $\tau$ between non-anchor prober ranks across anchor choices. Each cell compares the rank order of probers (by composition-gap count) under two different anchor screens, computed on the four probers the two anchor pairs share. Diagonal is $1$ by definition.}
\label{tab:anchor_kendall}
\vspace{4pt}
\begin{tabular}{lcccccc}
\toprule
\textbf{Anchor} & \textbf{Qwen-b.} & \textbf{Outc-2K} & \textbf{Mile-2K} & \textbf{gpt-oss} & \textbf{Llama-70B} & \textbf{DSeek} \\
\midrule
Qwen-b. & 1.00 & +0.67 & +0.33 & +0.33 & +1.00 & +0.33 \\
Outc-2K & +0.67 & 1.00 & +1.00 & +0.33 & +0.67 & +0.33 \\
Mile-2K & +0.33 & +1.00 & 1.00 & +0.67 & +0.67 & +0.67 \\
gpt-oss & +0.33 & +0.33 & +0.67 & 1.00 & +0.67 & +1.00 \\
Llama-70B & +1.00 & +0.67 & +0.67 & +0.67 & 1.00 & +0.67 \\
DSeek & +0.33 & +0.33 & +0.67 & +1.00 & +0.67 & 1.00 \\
\bottomrule
\end{tabular}
\end{table}

\paragraph{Recovered \textsc{Integrate} milestones (auxiliary release).} The pipeline that produced the diagnostic set kept the per-family \textsc{Integrate} type label in the released artifact but did not write the \textsc{Integrate} prompt or gold answer to the diagnostic file. We later re-ran the teacher in \textsc{decompose} mode to fill in those two fields. The recovered packets pass our schema check, but we do not claim that every recovered \textsc{Integrate} gold answer is correct: when the parent's gold answer is in an unusual format (free-form English, an answer-key string like \texttt{\textbackslash textbf\{(E)\}15}, or a placeholder like \texttt{\textbackslash text\{None\}}), the strict symbolic check between the teacher's answer and the parent's gold answer fails even when the teacher's output is mathematically defensible. We release these packets as a reading aid for auditors who want to inspect the teacher's intended synthesis step. They are not used in any main-paper number.

\begin{table}[h]
\centering
\small
\caption{Milestone types and compiled-milestone distribution across the diagnostic set.}
\label{tab:milestone_types}
\vspace{4pt}
\begin{tabular}{llr}
\toprule
\textbf{Type} & \textbf{Role} & \textbf{\%} \\
\midrule
\textsc{Key\_Move} & Critical algebraic/geometric step & 50.7 \\
\textsc{Model} & Problem modeling or formulation & 24.6 \\
\textsc{Compute} & Direct computation & 16.2 \\
\textsc{Normalize} & Standardize representation & 6.4 \\
\textsc{Other} & Lemma, Warmup, Sanity & 2.1 \\
\bottomrule
\end{tabular}
\end{table}

Typed taxonomies for reasoning steps have precedent in behavioral testing frameworks for NLP~\citep{ribeiro2020checklist} and in error taxonomies for math solutions~\citep{uesato2022solving}. Our types are chosen for their separability under symbolic verification rather than for cognitive realism. The anchor-solvable family definition is: (i) the parent is unsolved under $K{=}8$ screening rollouts under Qwen-base, (ii) every Stage~0-tested milestone solves at least once under the same screening, and (iii) at least one milestone is non-\textsc{Integrate}. Of 2{,}000 candidates, 731 satisfy this. We restrict to $\geq 2$ non-\textsc{Integrate} milestones for $n{=}354$. The panel results use the version of the set whose gold answers were repaired before the panel runs, and the repair log is released. Appendix~\ref{app:audit_rebuttal} lists ten further defects found afterwards.

\paragraph{Milestone-type breakdown by parent-probe outcome (descriptive).} Table~\ref{tab:milestone_types_by_state} reports, for each MilestoneRL-2K family state, the fraction of families whose milestone set contains at least one milestone of each type (rows are not mutually exclusive). \textsc{Model} milestones (problem formulation, setup) are over-represented in \textsc{Roadmap-Needed} and \textsc{Answers-Needed} states (57\% and 59\%) relative to \textsc{Direct} (44\%), while \textsc{Key\_Move} dominance is flat (82--86\%). We treat this breakdown as descriptive, since milestone types are teacher-assigned and not independently validated.

\begin{table}[h]
\centering
\footnotesize
\caption{Milestone-type composition of families in each MilestoneRL-2K state. Each cell is the percentage of families in that state whose milestone set contains at least one milestone of the given type (rows are not mutually exclusive, and a family can contribute to multiple columns).}
\label{tab:milestone_types_by_state}
\vspace{4pt}
\setlength{\tabcolsep}{5pt}
\begin{tabular}{lrrrrrr}
\toprule
\textbf{MilestoneRL-2K state} & \textbf{$n$} & \textbf{\textsc{Key\_Move}} & \textbf{\textsc{Model}} & \textbf{\textsc{Compute}} & \textbf{\textsc{Normalize}} & \textbf{Other} \\
\midrule
\textsc{Direct} & 96 & 86\% & 44\% & 27\% & 15\% & 6\% \\
\textsc{Roadmap-Needed} & 47 & 83\% & 57\% & 45\% & 6\% & 6\% \\
\textsc{Answers-Needed} & 22 & 82\% & 59\% & 45\% & 9\% & 5\% \\
\textsc{Unrecovered} & 189 & 86\% & 48\% & 32\% & 17\% & 3\% \\
\bottomrule
\end{tabular}
\end{table}

\section{Exact Oracle Prompts}
\label{app:prompts}

This section gives the exact student-side prompts used for the parent probes and matched control conditions. The same system prompt and the same boxed-answer instruction are used across all conditions. Only the additional context appended after the parent prompt varies.

The student is always wrapped with the system prompt: \textit{``Please reason step by step, and put your final answer within $\backslash$\texttt{boxed\{\}}. If the answer is a formula or number, use standard LaTeX notation\ldots''}

\paragraph{C1 (Direct).} The parent problem prompt only:
\begin{quote}\small\ttfamily
Solve the following problem. Provide your final answer clearly.

Problem: \{parent\_text\}
\end{quote}

\paragraph{C2 (Roadmap).} The parent prompt followed by:
\begin{quote}\small\ttfamily
Hint -{}-{}- the following sub-goals structure the solution:

Sub-goal 1: \{milestone\_1\_description\}

Sub-goal 2: \{milestone\_2\_description\}

\ldots
\end{quote}

\paragraph{C3 (Roadmap + milestone answers).} The parent prompt followed by:
\begin{quote}\small\ttfamily
The following sub-goals have been solved:

Sub-goal 1: \{milestone\_1\_description\}

\quad Answer: \{gold\_answer\_1\}

Sub-goal 2: \{milestone\_2\_description\}

\quad Answer: \{gold\_answer\_2\}

\ldots

Use these results to solve the original problem.
\end{quote}

For control conditions, the structure is identical but with content substituted:
\textbf{C2-random} replaces milestone descriptions with descriptions from a randomly selected different milestone family.
\textbf{C2-generic} replaces sub-goal descriptions with the verbatim generic four-step decomposition prompt: \textit{``Hint --- to solve this problem, consider the following approach: (1) Identify the key mathematical concepts and relationships involved. (2) Break the problem into smaller, manageable sub-steps. (3) Solve each sub-step carefully, checking your work. (4) Combine the results to reach the final answer.''} (in the spirit of least-to-most prompting~\citep{Zhou2022LeastToMost}).
\textbf{C3-mismatched} keeps the correct milestone descriptions but replaces the gold answers with answers sampled from other milestone families.

\section{Threshold Sensitivity Analysis}
\label{app:threshold}

We assess sensitivity to the default $\geq 1$-of-$K$ success threshold used to assign parent-probe outcomes in the main text. Table~\ref{tab:threshold} sweeps the threshold $\tau \in \{1,2,3,4,5\}$ across all three conditions simultaneously: a family is classified \textsc{Direct} iff $C_1 \geq \tau$, \textsc{Roadmap-Needed} iff $C_1 < \tau$ and $C_2 \geq \tau$, and analogously for \textsc{Answers-Needed} and \textsc{Unrecovered}.

\begin{table}[h]
\centering
\small
\caption{Family state counts as the threshold $\tau$ varies (MilestoneRL-2K, $n{=}354$). Row sum = 354 throughout. \textsc{Unrecovered} dominates for every $\tau$.}
\label{tab:threshold}
\vspace{4pt}
\begin{tabular}{ccccc}
\toprule
$\tau$ & \textsc{Direct} & \textsc{Roadmap-Needed} & \textsc{Answers-Needed} & \textsc{Unrecovered} \\
\midrule
1 (default) & 96 & 47 & 22 & 189 \\
2 & 65 & 44 & 19 & 226 \\
3 & 46 & 40 & 17 & 251 \\
4 & 36 & 27 & 21 & 270 \\
5 & 26 & 21 & 21 & 286 \\
\bottomrule
\end{tabular}
\end{table}

\textsc{Unrecovered} is the largest non-\textsc{Direct} bucket for every $\tau$, covering 53\% of all families at $\tau{=}1$ (default) and 81\% at $\tau{=}5$, while the \textsc{Roadmap-Needed} and \textsc{Answers-Needed} families together shrink from 69 at $\tau{=}1$ to 42 at $\tau{=}5$. The ordering of the states is the same at every threshold.

\section{Specificity Tables and Per-Contrast Interpretation}
\label{app:cross_model_stats}

This appendix gives full per-contrast counts and paired exact $p$-values for the specificity result reported in Section~\ref{sec:oracle_specificity} and the cross-model panel in Section~\ref{sec:cross_model}. All tests are paired exact McNemar~\citep{mcnemar1947} at the family level, $K{=}8$ rollouts per condition, \texttt{max\_tokens=16384} throughout.

\begin{table}[h]
\centering
\small
\caption{MilestoneRL-2K corruption controls on the $n{=}354$ main set. \textsc{sr} is per-rollout success rate. $C_2$-random and $C_2$-generic do not improve over direct prompting. $C_3$-mismatched lifts above $C_1$ but stays below $C_3$-gold (the gap reflects the additional value of problem-matched milestone answers). Only $C_2$-correct and $C_3$-gold reach substantial lift. (Section~\ref{sec:oracle_specificity} reports the headline numbers in prose.)}
\label{tab:ablation}
\vspace{4pt}
\begin{tabular}{lcccc}
\toprule
\textbf{Condition} & \textbf{Solved} & \textbf{sr} & \textbf{$\Delta$ vs $C_1$} & \textbf{paired $p$} \\
\midrule
$C_1$ direct & 96 & .106 & --- & --- \\
$C_2$-random & 78 & .072 & $-$18 & $0.010$ \\
$C_2$-generic & 96 & .090 & $0$ & $1.00$ \\
\textbf{$C_2$-correct} & \textbf{123} & \textbf{.145} & \textbf{+27} & $\boldsymbol{0.001}$ \\
$C_3$-mismatched & 119 & .137 & +23 & $0.011$ \\
\textbf{$C_3$-gold} & \textbf{140} & \textbf{.195} & \textbf{+44} & $\boldsymbol{3 \times 10^{-7}}$ \\
\bottomrule
\end{tabular}
\end{table}

\begin{table}[h]
\centering
\footnotesize
\caption{Per-model paired exact McNemar on the 354-family panel. Each cell: $\Delta$ (families the correct condition solves but the control does not, minus the reverse), two-sided exact $p$-value. Roadmap-specific lift is significant for all six models. $C_3$-gold vs.\ mismatched saturates for OutcomeRL-2K, gpt-oss-20b, and DeepSeek-V3.1.}
\label{tab:cross_model_specificity_stats}
\vspace{4pt}
\setlength{\tabcolsep}{4pt}
\begin{tabular}{lcccc}
\toprule
\textbf{Model} & \textbf{$C_2$-corr vs $C_1$} & \textbf{vs $C_2$-rand} & \textbf{vs $C_2$-gen} & \textbf{$C_3$-gold vs $C_3$-mism} \\
\midrule
Qwen-base & $+31$, $2.4\text{e-}4$ & $+51$, $8.6\text{e-}9$ & $+35$, $8.2\text{e-}5$ & $+18$, $0.015$ \\
OutcomeRL-2K step 180 & $+29$, $7.7\text{e-}4$ & $+28$, $7.6\text{e-}4$ & $+27$, $0.003$ & $+11$, $0.144$ \\
MilestoneRL-2K step 180 & $+27$, $0.001$ & $+45$, $2.4\text{e-}7$ & $+27$, $0.002$ & $+21$, $0.005$ \\
gpt-oss-20b & $+22$, $0.003$ & $+20$, $0.010$ & $+19$, $0.011$ & $+7$, $0.324$ \\
Llama-3.3-70B-Inst. & $+42$, $4.5\text{e-}10$ & $+46$, $3.2\text{e-}10$ & $+34$, $2\text{e-}6$ & $+27$, $4.6\text{e-}4$ \\
DeepSeek-V3.1 & $+26$, $4.2\text{e-}5$ & $+29$, $9\text{e-}6$ & $+17$, $0.006$ & $+3$, $0.701$ \\
\bottomrule
\end{tabular}
\end{table}

\begin{figure}[h]
\centering
\includegraphics[width=0.98\textwidth]{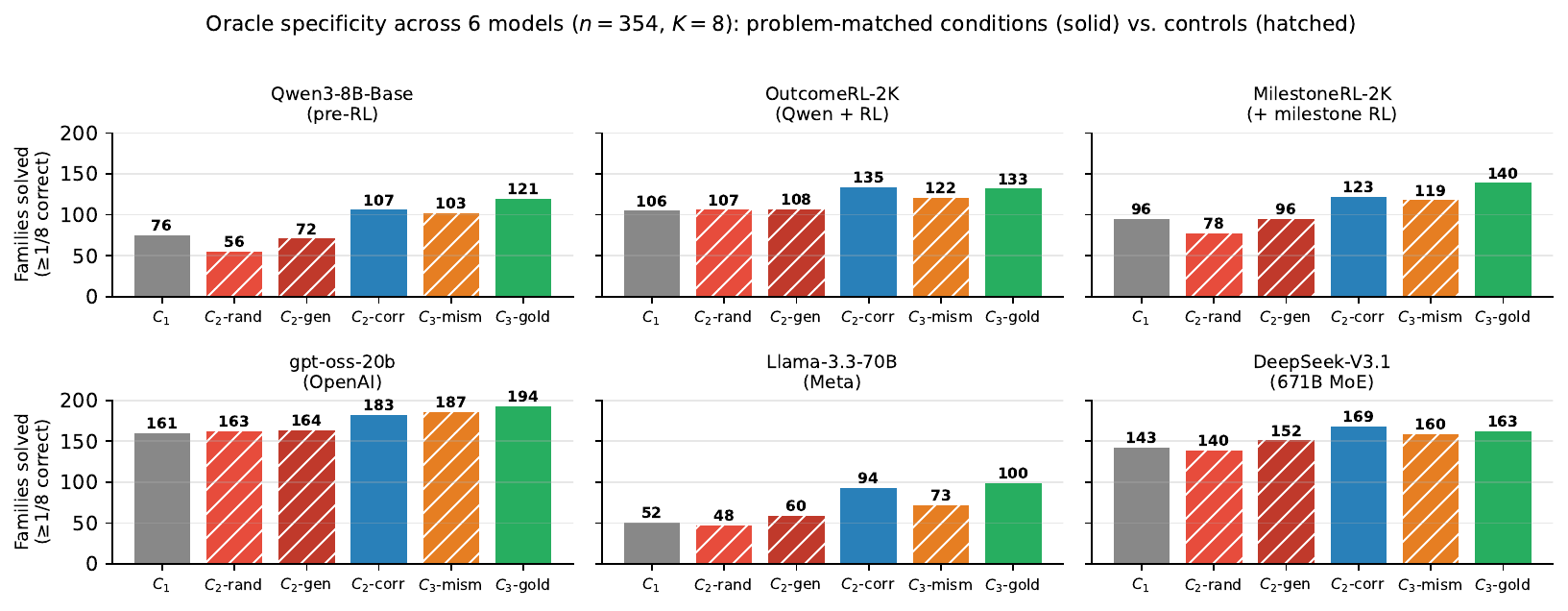}
\caption{Specificity across all six models on the $n{=}354$ diagnostic set. Solid bars are problem-matched conditions ($C_1$ direct in gray, $C_2$-correct in blue, $C_3$-gold in green). Hatched red bars are control conditions ($C_2$-random, $C_2$-generic, $C_3$-mismatched). Numbers above bars: families solved with $\geq 1/8$ correct rollouts. No control exceeds its matched correct counterpart. Per-model paired $p$-values are in Table~\ref{tab:cross_model_specificity_stats}.}
\label{fig:cross_model_specificity}
\end{figure}

\emph{Multiple-testing hygiene.} The four per-model contrasts are the planned comparisons fixed by the protocol's three oracle conditions and three corruption controls (Section~\ref{sec:method}), so we report uncorrected exact $p$-values. BH--FDR correction at $q{=}0.05$ across the 24 tests would not change which $C_2$ contrasts reach significance. The three non-significant $C_3$-gold vs.\ $C_3$-mismatched results (OutcomeRL-2K $p=0.144$, gpt-oss $p=0.324$, DeepSeek $p=0.701$) remain non-significant under any reasonable correction.

\emph{Why the $C_3$ contrast saturates on strong models.} At 16K, strong and reasoning-trained models can recover from a mismatched gold answer given enough tokens, so the $C_3$-gold vs.\ $C_3$-mismatched contrast loses statistical distinguishability on those models. The $C_2$ contrasts remain significant, indicating that problem-matched roadmap structure still separates from random or generic controls. The saturation pattern indicates that on stronger models the specific answer content adds less marginal lift once the model has room to reason.

\section{Bootstrap CIs and Pairwise Ranking Stability}
\label{app:bootstrap}

We quantify the stability of the cross-model any-probe ranking under a non-parametric family-level bootstrap (1{,}000 resamples of the 354 families with replacement, seed 42). Table~\ref{tab:bootstrap_cis} reports central tendency and 95\% percentile intervals, and Table~\ref{tab:bootstrap_ordering} reports pairwise ordering frequency on the any-probe count.

\begin{table}[h]
\centering
\footnotesize
\caption{Bootstrap 95\% CIs on fingerprint counts (1000 resamples of 354 families with replacement). \textsc{direct} count and any-probe count ($= \textsc{Direct}+\textsc{Roadmap-Needed}+\textsc{Answers-Needed}$) with central tendency and $[2.5, 97.5]$ percentile interval.}
\label{tab:bootstrap_cis}
\vspace{4pt}
\setlength{\tabcolsep}{6pt}
\begin{tabular}{lccc}
\toprule
\textbf{Model} & \textbf{\textsc{direct}} & \textbf{Any probe} & \textbf{Any-probe rank} \\
 & (mean [95\% CI]) & (mean [95\% CI]) & (95\% CI range) \\
\midrule
Qwen-base & 76.0 [61, 92] & 152.4 [134, 171] & 5 [4, 5] \\
OutcomeRL-2K & 106.0 [90, 123] & 170.3 [153, 189] & 3 [2, 4] \\
MilestoneRL-2K & 95.8 [80, 112] & 165.2 [147, 184] & 4 [3, 4] \\
gpt-oss-20b & 160.9 [143, 178] & 213.9 [196, 232] & 1 [1, 1] \\
Llama-70B & 51.8 [39, 64] & 126.0 [109, 145] & 6 [6, 6] \\
DeepSeek-V3.1 & 143.1 [125, 160] & 183.0 [165, 200] & 2 [2, 4] \\
\bottomrule
\end{tabular}
\end{table}

\begin{table}[h]
\centering
\scriptsize
\caption{Pairwise ordering frequency on the any-probe count: fraction of 1000 bootstrap resamples (family-level) where the row model's any-probe count exceeds the column model's. Values near 0.5 indicate unstable order, and values near 1.0 indicate stable dominance.}
\label{tab:bootstrap_ordering}
\vspace{4pt}
\setlength{\tabcolsep}{4pt}
\begin{tabular}{lcccccc}
\toprule
& Qwen-base & OutcomeRL-2K & MilestoneRL-2K & gpt-oss-20b & Llama-70B & DeepSeek-V3.1 \\
\midrule
Qwen-base & --- & 0.01 & 0.02 & 0.00 & 1.00 & 0.00 \\
OutcomeRL-2K & 0.99 & --- & 0.81 & 0.00 & 1.00 & 0.10 \\
MilestoneRL-2K & 0.97 & 0.15 & --- & 0.00 & 1.00 & 0.03 \\
gpt-oss-20b & 1.00 & 1.00 & 1.00 & --- & 1.00 & 1.00 \\
Llama-70B & 0.00 & 0.00 & 0.00 & 0.00 & --- & 0.00 \\
DeepSeek-V3.1 & 1.00 & 0.89 & 0.96 & 0.00 & 1.00 & --- \\
\bottomrule
\end{tabular}
\end{table}

Cross-lineage orderings are stable: gpt-oss-20b dominates and Llama-3.3-70B is dominated in 100\% of resamples, and DeepSeek-V3.1 exceeds the three Qwen3-8B checkpoints in 89--100\%. The within-Qwen RL gain over Qwen-base is also stable (99\% for OutcomeRL-2K, 97\% for MilestoneRL-2K). OutcomeRL-2K versus MilestoneRL-2K is not clearly separated on any-probe count alone (OutcomeRL-2K above MilestoneRL-2K in 81\%). This is why the longitudinal claim in Section~\ref{sec:results_training} focuses on the shape of redistribution rather than the any-probe ranking.

\section{Stage 0 Milestone Solvability and Milestone-Test-Pass Fingerprints}
\label{app:stage0}

This appendix reports per-model Stage 0 milestone-test statistics, which feed the reasoning-gap taxonomy in Section~\ref{sec:lattice}. Stage 0 is the milestone-only test: a family \emph{passes the milestone test} for a model iff that model solves every non-\textsc{Integrate} milestone with $\geq 1/8$ correct rollouts in isolation. Stage 0 and parent probing both run at \texttt{max\_tokens=16384} and $K{=}8$. Per-milestone solve rates are high for every model (87\%--95\%), so most unrecovered cases are not explained by missing milestone-level solvability alone. Table~\ref{tab:stage0} reports the milestone-test-pass count and the family-state fingerprint on the all-354 set and the test-pass subset.

\begin{table}[h]
\centering
\footnotesize
\caption{Stage 0 milestone solvability at \texttt{max\_tokens=16384}, $K{=}8$. A family \emph{passes the milestone test} iff the model solves every non-\textsc{Integrate} milestone with $\geq 1/8$ correct rollouts. Fingerprint cells: \textsc{Direct}/\textsc{Roadmap-Needed}/\textsc{Answers-Needed}/\textsc{Unrecovered}.}
\label{tab:stage0}
\vspace{4pt}
\begin{tabular}{lcccc}
\toprule
\textbf{Model} & \textbf{Milestone solve rate} & \textbf{Passing families} & \textbf{All families} & \textbf{Passing families only} \\
\midrule
Qwen-base & 93\% & 306 & 76/50/26/202 & 70/42/23/171 \\
OutcomeRL-2K step 180 & 92\% & 296 & 106/50/14/184 & 88/44/9/155 \\
MilestoneRL-2K step 180 & 95\% & 314 & 96/47/22/189 & 88/40/19/167 \\
gpt-oss-20b & 93\% & 306 & 161/36/17/140 & 141/33/14/118 \\
Llama-3.3-70B-Inst. & 87\% & 268 & 52/46/28/228 & 38/38/21/171 \\
DeepSeek-V3.1 & \textbf{87\%} & \textbf{264} & 143/33/7/171 & 108/25/6/125 \\
\bottomrule
\end{tabular}
\end{table}

Milestone solve rate $=$ per-milestone solve rate ($\geq 1/8$) across all non-\textsc{Integrate} milestones. Families passing test $=$ families for which every non-\textsc{Integrate} milestone has $\geq 1/8$ correct. Fingerprint cells count \textsc{Direct}/\textsc{Roadmap-Needed}/\textsc{Answers-Needed}/\textsc{Unrecovered} on each denominator.

\paragraph{Stage 0 at the 4K screening budget.} The diagnostic-set construction step was originally run at the 4K decoding budget. We re-ran Stage 0 at the 16K parent-probing budget for the main results above, and report the 4K-vs-16K deltas in Table~\ref{tab:stage0_budget}. Per-model milestone-test pass counts shift by at most 8 families, and composition-gap counts shift by at most 7, so the Stage 0 reading is stable across the two budgets.

\begin{table}[h]
\centering
\footnotesize
\caption{Stage 0 replication at the 16K decoding budget. Pass counts are families that pass the milestone test (every non-\textsc{Integrate} milestone solves at least once at $K{=}8$). Composition-gap counts are families that pass Stage 0 and are \textsc{Unrecovered} under all three parent probes.}
\label{tab:stage0_budget}
\vspace{4pt}
\begin{tabular}{lccccc}
\toprule
\textbf{Model} & \textbf{Pass (4K)} & \textbf{Pass (16K)} & \textbf{Pass flipped} & \textbf{Comp.\ gap (4K)} & \textbf{Comp.\ gap (16K)} \\
\midrule
Qwen-base & 303 & 306 & 43 & 173 & 171 \\
OutcomeRL-2K & 300 & 296 & 38 & 155 & 155 \\
MilestoneRL-2K & 308 & 314 & 36 & 167 & 167 \\
gpt-oss-20b & 304 & 306 & 22 & 120 & 118 \\
Llama-70B & 261 & 268 & 39 & 164 & 171 \\
DeepSeek-V3.1 & 256 & 264 & 60 & 127 & 125 \\
\bottomrule
\end{tabular}
\end{table}

\begin{table}[h]
\centering
\small
\caption{Reasoning-gap taxonomy: Stage 0 milestone test (pass/fail) $\times$ smallest help needed to unlock the parent, per model on the 354-family diagnostic set. Counts are families. Both Stage 0 and parent probing at \texttt{max\_tokens=16384}, $K{=}8$.}
\label{tab:bottleneck_lattice}
\vspace{4pt}
\setlength{\tabcolsep}{4pt}
\begin{tabular}{lrrrrrr}
\toprule
\textbf{Model} & \textbf{Direct} & \textbf{Roadmap} & \textbf{MS-Exec} & \textbf{Composition} & \textbf{Missing-MS} & \textbf{Capability} \\
\midrule
Qwen-base & 76 & 42 & 23 & 171 & 11 & 31 \\
OutcomeRL-2K & 106 & 44 & 9 & 155 & 11 & 29 \\
MilestoneRL-2K & 96 & 40 & 19 & 167 & 10 & 22 \\
gpt-oss-20b & 161 & 33 & 14 & 118 & 6 & 22 \\
Llama-70B & 52 & 38 & 21 & 171 & 15 & 57 \\
DeepSeek-V3.1 & 143 & 25 & 6 & 125 & 9 & 46 \\
\bottomrule
\end{tabular}
\end{table}

\section{Aggregate-Comparator Evaluation Settings and External Reference Table}
\label{app:aggregate_settings}
\label{app:external_aggregate}

This appendix documents the MATH hard-100 external comparator and reports the joint MATH-vs-fingerprint table referenced from Section~\ref{sec:cross_model}. MATH hard-100 is a different problem distribution from the 354-family diagnostic set, so it is supporting context, not the primary comparison. Table~\ref{tab:same_set_summary} is the same-set version.

\begin{table}[t]
\centering
\footnotesize
\caption{\textbf{External MATH score and diagnostic fingerprint.}
MATH hard-100 provides an external scalar comparator. The fingerprint columns summarize the 354-family diagnostic set.}
\label{tab:aggregate_vs_fingerprint}
\vspace{3pt}
\setlength{\tabcolsep}{3.5pt}
\begin{tabular}{@{}lrrrrrrr@{}}
\toprule
& \multicolumn{2}{c}{\textbf{MATH hard-100}} 
& \multicolumn{5}{c}{\textbf{Diagnostic fingerprint, $n{=}354$}} \\
\cmidrule(lr){2-3} \cmidrule(lr){4-8}
\textbf{Model} 
& \textbf{Score}
& \textbf{p@8}
& \textbf{Direct} 
& \textbf{Roadmap} 
& \textbf{Answers} 
& \textbf{Unrec.} 
& \shortstack{\textbf{Any}\\\textbf{probe}} \\
\midrule
Qwen-base        & 0.45 & 0.75 &  76 & 50 & 26 & 202 & 152 \\
OutcomeRL-2K     & 0.65 & 0.88 & 106 & 50 & 14 & 184 & 170 \\
MilestoneRL-2K   & 0.63 & 0.85 &  96 & 47 & 22 & 189 & 165 \\
gpt-oss-20b      & 0.89 & 0.97 & 161 & 36 & 17 & 140 & 214 \\
Llama-70B        & 0.49 & 0.68 &  52 & 46 & 28 & 228 & 126 \\
DeepSeek-V3.1    & 0.85 & 0.96 & 143 & 33 &  7 & 171 & 183 \\
\bottomrule
\end{tabular}

\vspace{3pt}
\parbox{0.96\linewidth}{\footnotesize
\emph{Notes.} MATH score is mean single-rollout correctness estimated from $K{=}8$ samples on the 100-problem MATH hard-100 subset, and p@8 is the $\geq$1-of-8 solve rate. Diagnostic columns report family-level counts on the 354 milestone families. Any probe $=$ Direct $+$ Roadmap $+$ Answers. All evaluations use \texttt{max\_tokens=16384} and the same symbolic verifier.
}
\end{table}

\paragraph{Evaluation set.} All models use the same 100 level-5 problems from the MATH500 test split~\citep{hendrycksmath2021,lightman2023verify}. The 4K and 16K panel outputs are released as \texttt{math\_hard100\_panel.jsonl} and \texttt{math\_hard100\_panel\_16k.jsonl}. The raw MATH problems themselves are part of the public MATH benchmark and are referenced by problem ID rather than re-released.

\paragraph{Prompt and decoding.} The student system prompt (Appendix~\ref{app:prompts}) is identical for all models. Decoding: temperature 1.0, \texttt{max\_tokens}${=}$16384 per rollout (matching the main diagnostic panel). Each model is rendered with its native chat template. For Qwen-family models we use the \texttt{role\_colon} renderer that was active during training, for Llama-3.3-70B-Instruct the \texttt{llama3} renderer, and for gpt-oss-20b / DeepSeek-V3.1 the \texttt{role\_colon} renderer (no dedicated chat template required for these checkpoints). Answer extraction and verification are the same symbolic cascade used in the oracle (Appendix~\ref{app:verifier_ext}).

\paragraph{pass@1 and pass@8.} All six models are evaluated with the same $K{=}8$ rollouts on the same 100 problems, using the same prompt, decoding, and symbolic verifier. pass@1 in Table~\ref{tab:aggregate_vs_fingerprint} is mean single-rollout correctness $\hat p = \tfrac{1}{NK}\sum_{i,k}[\text{correct}]$, and pass@8 is the $\geq$1-of-8 solve rate on the same sample.

\paragraph{Formatting caveat.} Model-specific formatting and template defaults still affect surface extraction: a model that omits \texttt{\textbackslash boxed\{\}} may lose credit if the final answer is not extractable, even when the reasoning is correct. Appendices~\ref{app:audit_rebuttal} and~\ref{app:frontier_judge} measure the resulting verifier and format errors.

\section{Training Dynamics: Endpoint Numbers and Trajectory Tomography}
\label{app:training_dynamics}
\label{app:trajectory}

This appendix supports the longitudinal readout in Section~\ref{sec:results_training} with the exact two-arm endpoint numbers behind Figure~\ref{fig:two_arm_redistribution} and a per-checkpoint trajectory on intermediate Tinker checkpoints. Endpoints are at \texttt{max\_tokens=16384}, $K{=}8$, on the same 354-family diagnostic set. Both RL arms start from Qwen-base. The 4K wandb training-time validation curve is in Appendix~\ref{app:decoding_budget} alongside the 4K-vs-16K decoding-budget audit. Both RL arms transit through configurations whose any-probe count is below the Qwen-base starting point before recovering at the step-180 endpoint. This is a temporary drop in any-probe count that aggregate accuracy hides.

\begin{table}[h]
\centering
\small
\caption{Numerical values behind Figure~\ref{fig:two_arm_redistribution} (Section~\ref{sec:results_training}): two-arm longitudinal readout on the 354-family diagnostic set at \texttt{max\_tokens=16384}, $K{=}8$, starting from the same Qwen-base checkpoint. MATH hard-100 pass@1 shows both RL arms gaining comparably. The fingerprint shows them moving differently.}
\label{tab:two_arm_longitudinal}
\vspace{4pt}
\setlength{\tabcolsep}{4pt}
\resizebox{\linewidth}{!}{%
\begin{tabular}{lccccc}
\toprule
& \textbf{MATH score} & \textsc{Direct} & \textsc{Roadmap-Needed} & \textsc{Answers-Needed} & \textbf{Any probe} \\
\midrule
Qwen-base & 0.454 & 76 & 50 & 26 & 152 \\
OutcomeRL-2K step 180 & 0.654 ($+20.0$) & 106 ($+30$) & 50 ($+0$) & 14 ($-12$) & 170 ($+18$) \\
MilestoneRL-2K step 180 & 0.634 ($+18.0$) & 96 ($+20$) & 47 ($-3$) & 22 ($-4$) & 165 ($+13$) \\
\bottomrule
\end{tabular}}
\end{table}

\begin{figure}[h]
\centering
\includegraphics[width=\textwidth]{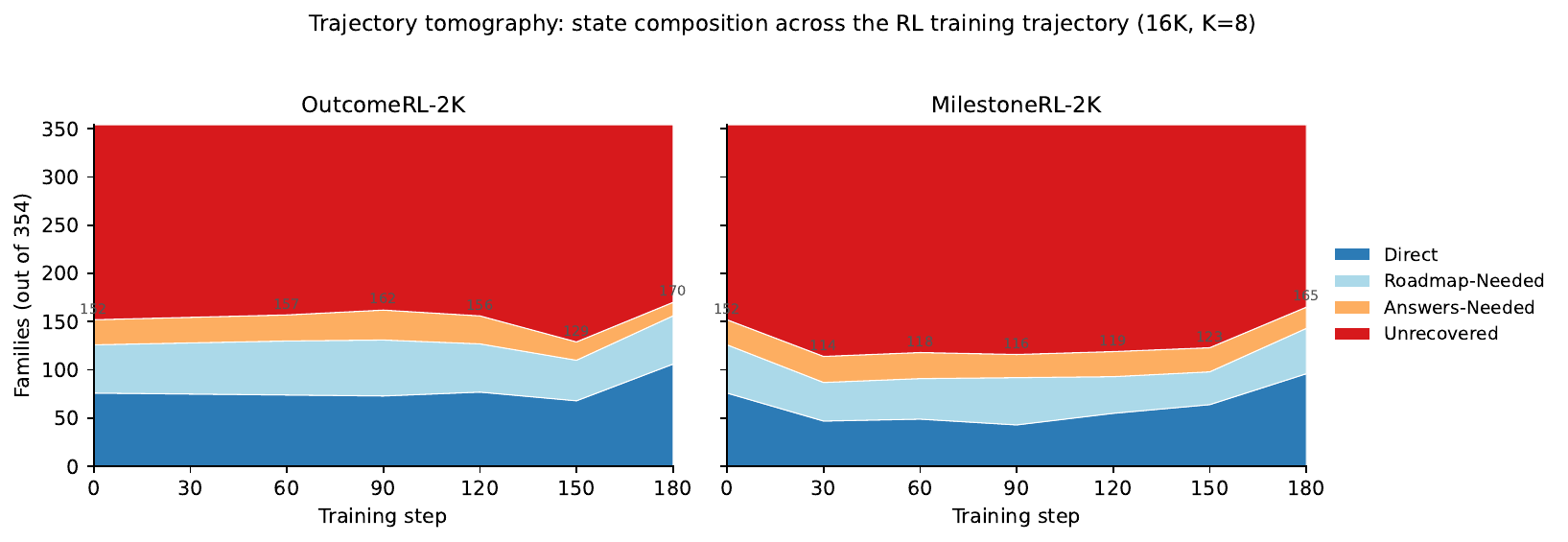}
\caption{\textbf{Training trajectories move through different diagnostic states.} Stacked areas show parent-probe outcomes across checkpoints. Numbers above each bar are families solved by any probe. The MATH hard-100 pass@1 trajectory over the same window is in Appendix~\ref{app:decoding_budget}.}
\label{fig:trajectory_tomography}
\end{figure}

\paragraph{Trajectory tomography.} We re-ran the full diagnostic ($K{=}8$, \texttt{max\_tokens=16384}, 354 families, six conditions) on intermediate checkpoints in addition to the Qwen-base and step-180 endpoints used in Section~\ref{sec:results_training}: \{30, 60, 90, 120, 150\} for MilestoneRL-2K (5 panels) and \{60, 90, 120, 150\} for OutcomeRL-2K (4 panels, because step 30 was not retained as a Tinker checkpoint). Each (arm, step) is a separate panel costing $354 \times 6 \times 8 = 16{,}992$ rollouts. The nine new panels add $\sim$153K rollouts on top of the main panel. Figure~\ref{fig:trajectory_tomography} shows the resulting state-composition trajectories. Two patterns are visible that the endpoint-only reading cannot expose. (i) MilestoneRL-2K depresses the any-probe count early ($152 \to 114$ at step 30, a $-38$-family drop) and remains $\sim 30$ families below the Qwen-base any-probe count for $\sim 120$ steps before recovering to 165 at step 180. (ii) OutcomeRL-2K stays close to baseline through step 120 (152, 157, 162, 156), dips at step 150 to 129, then jumps to 170 at step 180. Both arms therefore transit through configurations whose any-probe count is \emph{below the Qwen-base starting point}, even though MATH hard-100 pass@1 trends upward over the same window (training-time validation curve in Appendix~\ref{app:decoding_budget}, which is non-monotonic at the per-step level but ends well above the start). The fingerprint resolves a transient mid-training regression that aggregate accuracy hides. This is consistent with the cross-model finding that scalar accuracy and the any-probe count decouple, and adds that the same decoupling appears \emph{within} a single training run.

\section{RL Training-Recipe Disclosure for OutcomeRL-2K and MilestoneRL-2K}
\label{app:training_configs}

This appendix discloses which training-recipe dimensions are shared and which differ between OutcomeRL-2K and MilestoneRL-2K. Both endpoints initialize from the same Qwen-base checkpoint (Qwen/Qwen3-8B-Base, LoRA rank 64) and use the same GRPO recipe~\citep{shao2024deepseekmath} with DAPO-style dynamic sampling~\citep{yu2025dapo}, AdamW at LR $1{\times}10^{-5}$, KL coefficient $0$, group size 16, decoding $\texttt{max\_tokens}{=}16384$ at temperature $1.0$, the same strict symbolic-cascade in-loop verifier, the same 2{,}000-parent NuminaMath training pool (disjoint from the 354 diagnostic families), and the same 180-step horizon. The two arms differ on multiple dimensions beyond reward construction. The pair is therefore a longitudinal diagnostic case study, not a controlled ablation of milestone reward. Per-dimension training-row counts, the SFT warmup R1-distilled milestone CoT data, and the curriculum-filter source code are included with the artifact under \texttt{code/decomposer/pipeline/}. An exhaustive hyperparameter dump (every flag value used at training time) is not part of this release.

\begin{table}[h]
\centering
\small
\caption{Training-recipe dimensions of OutcomeRL-2K and MilestoneRL-2K. The two runs are not matched in compute.}
\label{tab:training_configs}
\vspace{4pt}
\setlength{\tabcolsep}{4pt}
\begin{tabular}{p{0.27\linewidth} p{0.13\linewidth} p{0.50\linewidth}}
\toprule
\textbf{Dimension} & \textbf{Shared / different?} & \textbf{Summary} \\
\midrule
Base model, optimizer, algorithm, horizon & shared & Qwen-base, GRPO, AdamW LR $1{\times}10^{-5}$, 180 steps \\
Reward signal & different & parent correctness only vs.\ parent + milestone rollouts \\
Per-step rollout budget & different & $1{,}024$ vs.\ $512$ ($\sim$184K vs.\ $\sim$92K total rollouts) \\
SFT warmup & different & none vs.\ 3-epoch R1-distilled CoT warmup \\
Training-data composition / curriculum & different & details in artifact (originals only vs.\ originals + milestones, distinct $\hat p$ filters) \\
\bottomrule
\end{tabular}
\end{table}

Both arms reach near-identical aggregate accuracy on MATH hard-100 ($0.654$ and $0.634$ pass@1). The diagnostic surfaces fingerprint differences that the aggregate hides.

\section{Cross-Model Panel Details}
\label{app:panel}

\begin{figure}[h]
\centering
\includegraphics[width=0.88\textwidth]{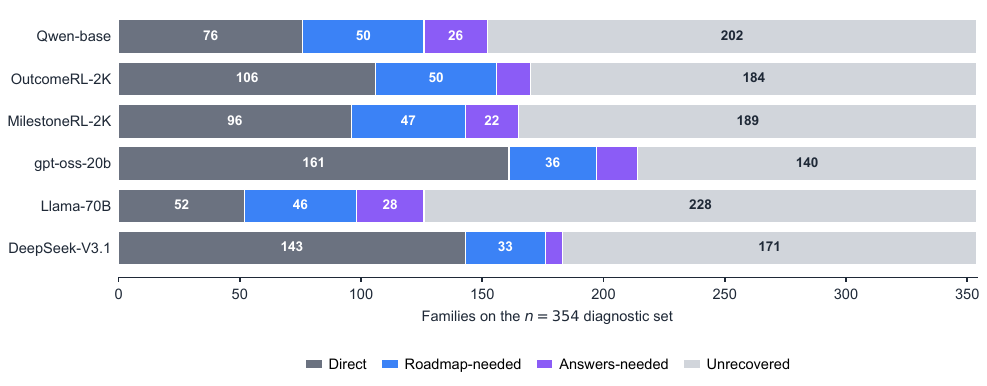}
\caption{Cross-model parent-probe outcome fingerprints on the $n{=}354$ diagnostic set. Each bar partitions families by the smallest help level that unlocks the parent: \textsc{Direct} ($C_1$), \textsc{Roadmap-Needed} ($C_2$), \textsc{Answers-Needed} ($C_3$), or \textsc{Unrecovered}. The aggregate counts behind each bar are reported in Table~\ref{tab:panel_details}.}
\label{fig:panel_fingerprints}
\end{figure}

This appendix reports per-condition solve counts and panel metadata for the six model instances in Section~\ref{sec:cross_model} on the same 354-family diagnostic set with $K{=}8$ rollouts per condition at \texttt{max\_tokens=16384}. Table~\ref{tab:panel_models} lists panel metadata, Table~\ref{tab:panel_details} gives the full per-condition solve counts, and Figure~\ref{fig:panel_fingerprints} visualizes the four-state parent-probe outcome composition.

\begin{table}[h]
\centering
\small
\caption{Cross-model panel metadata. OutcomeRL-2K and MilestoneRL-2K are RL-trained from the Qwen-base checkpoint sharing base, LoRA rank 64, GRPO recipe, 180-step training horizon, and the 2{,}000-parent NuminaMath training pool. They differ in reward construction (outcome-only versus milestone-augmented) and in the other training details listed in Table~\ref{tab:training_configs}.}
\label{tab:panel_models}
\vspace{4pt}
\begin{tabular}{llllr}
\toprule
\textbf{Model} & \textbf{Lineage} & \textbf{Parameters} & \textbf{Architecture} & \textbf{Training regime} \\
\midrule
Qwen-base & Qwen3 & 8B & dense & base \\
OutcomeRL-2K & Qwen3 & 8B & dense & RL-tuned (outcome) \\
MilestoneRL-2K & Qwen3 & 8B & dense & RL-tuned (milestone) \\
gpt-oss-20b & gpt-oss & 20B (3.6B active) & MoE & reasoning-trained \\
Llama-3.3-70B-Instruct & Llama-3 & 70B & dense & instruction-tuned \\
DeepSeek-V3.1 & DeepSeek-V3 & 671B (37B active) & MoE & reasoning-trained \\
\bottomrule
\end{tabular}
\end{table}

\paragraph{Serving infrastructure.} The Qwen-lineage panels (Qwen-base, OutcomeRL-2K, MilestoneRL-2K) are served via the Tinker managed-inference API. The same Tinker stack hosts the RL training for the two RL endpoints. The remaining three panel models (gpt-oss-20b, Llama-3.3-70B-Instruct, DeepSeek-V3.1) are accessed via hosted-inference APIs. The teacher model used during diagnostic-set construction (GPT-5.4 Thinking) and the R1 distillation pass also run on hosted inference. No local GPU cluster is required to reproduce the panel.

\begin{table}[h]
\centering
\footnotesize
\caption{Cross-model panel: families solved ($\geq 1$ correct rollout out of $K{=}8$) on the 354-family diagnostic set. Any probe $= \textsc{Direct} + \textsc{Roadmap-Needed} + \textsc{Answers-Needed}$ (out of 354) is the count of families solved by any probe.}
\label{tab:panel_details}
\vspace{4pt}
\begin{tabular}{lccccccc}
\toprule
& $C_1$ & $C_2$-rand & $C_2$-gen & $C_2$-corr & $C_3$-mism & $C_3$-gold & Any probe \\
\midrule
Qwen-base & 76 & 56 & 72 & 107 & 103 & 121 & 152 \\
OutcomeRL-2K step 180 & 106 & 107 & 108 & 135 & 122 & 133 & 170 \\
MilestoneRL-2K step 180 & 96 & 78 & 96 & 123 & 119 & 140 & 165 \\
gpt-oss-20b & 161 & 163 & 164 & 183 & 187 & 194 & 214 \\
Llama-3.3-70B-Instruct & 52 & 48 & 60 & 94 & 73 & 100 & 126 \\
DeepSeek-V3.1 & 143 & 140 & 152 & 169 & 160 & 163 & 183 \\
\bottomrule
\end{tabular}
\end{table}

\section{Decoding-Budget Sensitivity and Format Audit}
\label{app:decoding_budget}

This appendix documents the decoding-budget audit and shows what the panel looks like at the 4K budget used in early prototyping, alongside the 16K main-evaluation budget. Same 354-family panel, same prompts, same verifier. Only the decoding cap differs. We also include the 4K wandb training-time validation curve here, because its interpretation depends on the same 4K-vs-16K comparison.

Moving from 4K to 16K raises the any-probe count by 14 to 42 families for OutcomeRL-2K, gpt-oss-20b, and DeepSeek-V3.1, and changes it by at most 7 families for Qwen-base, MilestoneRL-2K, and Llama-3.3-70B-Instruct. Within the four outcomes, five of the six models move families mainly from \textsc{Answers-Needed} to \textsc{Direct}.

\begin{table}[h]
\centering
\footnotesize
\caption{\textbf{Decoding-budget sensitivity:} parent-probe outcome counts at \texttt{max\_tokens=4096} vs.\ \texttt{16384} on the same 354 families. OutcomeRL-2K, gpt-oss-20b, and DeepSeek-V3.1 gain 14 to 42 any-probe families at 16K. Qwen-base, MilestoneRL-2K, and Llama-3.3-70B-Instruct change by at most 7.}
\label{tab:panel_4k_vs_16k}
\vspace{4pt}
\setlength{\tabcolsep}{4pt}
\begin{tabular}{lrrrrrrrr}
\toprule
& \multicolumn{4}{c}{\textbf{4K}} & \multicolumn{4}{c}{\textbf{16K}} \\
\cmidrule(lr){2-5} \cmidrule(lr){6-9}
\textbf{Model} & \textsc{Direct} & \textsc{R-Need} & \textsc{A-Need} & \textbf{Any} & \textsc{Direct} & \textsc{R-Need} & \textsc{A-Need} & \textbf{Any} \\
\midrule
Qwen-base & 64 & 54 & 40 & 158 & 76 & 50 & 26 & 152 \\
OutcomeRL-2K & 79 & 50 & 27 & 156 & 106 & 50 & 14 & 170 \\
MilestoneRL-2K & 88 & 45 & 25 & 158 & 96 & 47 & 22 & 165 \\
gpt-oss-20b & 114 & 34 & 27 & 175 & 161 & 36 & 17 & 214 \\
Llama-70B & 54 & 48 & 26 & 128 & 52 & 46 & 28 & 126 \\
DeepSeek-V3.1 & 85 & 40 & 16 & 141 & 143 & 33 & 7 & 183 \\
\bottomrule
\end{tabular}
\end{table}

\paragraph{4K wandb training-time validation curve.} Table~\ref{tab:training_dynamics} reports the 13-checkpoint MATH hard-100 single-rollout pass@1 trajectory recorded at the 4K training-time validation budget. Endpoints read 0.41 (Qwen-base) and 0.62 (step 180) at 4K with $K{=}1$, against 0.454 and 0.654 at 16K with $K{=}8$ (Table~\ref{tab:two_arm_longitudinal}). The two budgets agree on the shape of the curve, and 16K is higher at both ends by 3 to 4 points.

\begin{table}[h]
\centering
\scriptsize
\caption{OutcomeRL-2K wandb single-rollout ($K{=}1$) MATH hard-100 pass@1 at 13 checkpoints, 15-step intervals, recorded at \texttt{max\_tokens=4096} (training-time validation budget, not the main parent-probing budget).}
\label{tab:training_dynamics}
\vspace{4pt}
\setlength{\tabcolsep}{4pt}
\begin{tabular}{lccccccccccccc}
\toprule
Step & 0 & 15 & 30 & 45 & 60 & 75 & 90 & 105 & 120 & 135 & 150 & 165 & 180 \\
\midrule
MATH score & 0.41 & 0.42 & 0.49 & 0.43 & 0.43 & 0.53 & 0.51 & 0.56 & 0.56 & 0.60 & 0.58 & 0.58 & 0.62 \\
\bottomrule
\end{tabular}
\end{table}

\paragraph{Format audit.} To confirm that the 4K shift is a decoding-budget artifact rather than a model property, we ran a focused audit on a 50-family random subset (seed 42) with four models (Qwen-base, MilestoneRL-2K, Llama-3.3-70B, DeepSeek-V3.1), conditions $C_1$ and $C_3$-gold, and $K{=}8$. Each rollout is classified into \textsc{no\_boxed\_trunc} (no boxed answer, response hit the 4K cap), \textsc{no\_boxed} (no boxed but not truncated), \textsc{unparsable}, \textsc{parsed\_wrong}, or \textsc{parsed\_correct}.

\begin{table}[h]
\centering
\scriptsize
\caption{\textbf{Format / extraction-failure audit on a 50-family random subset.} Percentages are of 400 rollouts per (model, condition) cell.}
\label{tab:format_audit}
\vspace{4pt}
\setlength{\tabcolsep}{4pt}
\begin{tabular}{llrrrrrr}
\toprule
\textbf{Model} & \textbf{Condition} & \textbf{$n$} & \textbf{no\_boxed} & \textbf{no\_boxed\_trunc} & \textbf{unparsable} & \textbf{parsed\_wrong} & \textbf{parsed\_correct} \\
\midrule
Qwen-base & $C_1$ & 400 & 13.0\% & 2.5\% & 0.0\% & 80.0\% & 4.5\% \\
Qwen-base & $C_3$-gold & 400 & 17.0\% & 3.8\% & 0.2\% & 62.3\% & 16.8\% \\
MilestoneRL-2K & $C_1$ & 400 & 8.5\% & 32.5\% & 0.0\% & 51.0\% & 8.0\% \\
MilestoneRL-2K & $C_3$-gold & 400 & 17.8\% & 12.5\% & 0.2\% & 51.0\% & 18.5\% \\
Llama-3.3-70B-Inst. & $C_1$ & 400 & 7.2\% & 0.5\% & 0.0\% & 88.2\% & 4.0\% \\
Llama-3.3-70B-Inst. & $C_3$-gold & 400 & 13.8\% & 0.0\% & 0.0\% & 71.0\% & 15.2\% \\
DeepSeek-V3.1 & $C_1$ & 400 & 8.0\% & 53.8\% & 0.5\% & 18.8\% & 19.0\% \\
DeepSeek-V3.1 & $C_3$-gold & 400 & 7.2\% & 44.0\% & 0.5\% & 24.8\% & 23.5\% \\
\bottomrule
\end{tabular}

\vspace{2pt}
\parbox{\linewidth}{\footnotesize \emph{Buckets.} \textsc{no\_boxed}: no \texttt{\textbackslash boxed\{\}} in response. \textsc{no\_boxed\_trunc}: no boxed and hit \texttt{max\_tokens}. \textsc{unparsable}: boxed but content unparseable. \textsc{parsed\_wrong}: parsed, verifier says wrong. \textsc{parsed\_correct}: parsed, verifier accepts.}
\end{table}

DeepSeek-V3.1 truncates on 53.8\% of $C_1$ rollouts and 44\% of $C_3$-gold rollouts at 4K. MilestoneRL-2K truncates 32.5\% at $C_1$. Llama-3.3-70B truncates 0\% but has elevated \textsc{parsed\_wrong} (88\% at $C_1$). \textsc{unparsable} is $\leq 0.5\%$ everywhere.

\paragraph{Consequences for main-text claims.} (i) At 4K, DeepSeek-V3.1 ranks 5th on the any-probe count (141/354), below all three Qwen3-8B checkpoints. At 16K it ranks 2nd (183/354), above every Qwen checkpoint. The apparent ``DeepSeek ranks 2nd on aggregate but 5th on any-probe count'' disagreement is a 4K artifact. (ii) At 4K, the Qwen training trajectory shows a near-flat any-probe count (Qwen-base 158, OutcomeRL-2K 156, MilestoneRL-2K 158). At 16K it shows a clear expansion (152 $\to$ 170 $\to$ 165). Reading the trajectory at 4K would lead to a different interpretation of RL training than 16K does. We report 16K throughout as the main evaluation budget.

\section{Held-out Replication Table}
\label{app:heldout}

This appendix replicates the specificity result on a 32-family held-out slice compiled from a 150-parent NuminaMath sample disjoint from the 2{,}000-parent main set, using the identical typed pipeline and verification cascade. The pattern reproduces: random and generic controls do not lift above $C_1$, while $C_2$-correct and $C_3$-gold do. $C_3$-mismatched also lifts ($p{=}0.031$) but stays below $C_3$-gold, mirroring the main set.

\begin{table}[h]
\centering
\small
\caption{Held-out replication ($n{=}32$, MilestoneRL-2K step 180). \textsc{sr} is per-rollout success rate. Random and generic controls do not lift above $C_1$. Correct roadmap and roadmap-plus-answers do. The mismatched-answer control lifts above $C_1$ but stays below $C_3$-gold.}
\label{tab:heldout}
\vspace{4pt}
\begin{tabular}{lcccc}
\toprule
\textbf{Condition} & \textbf{Solved} & \textbf{sr} & \textbf{$\Delta$ vs $C_1$} & \textbf{paired exact $p$} \\
\midrule
$C_1$ direct & 13 & .191 & --- & --- \\
$C_2$-random & 12 & .176 & $-$1 & $1.00$ \\
$C_2$-generic & 13 & .199 & 0 & $1.00$ \\
\textbf{$C_2$-correct} & \textbf{20} & \textbf{.438} & \textbf{+7} & $\boldsymbol{0.016}$ \\
$C_3$-mismatched & 19 & .426 & +6 & $0.031$ \\
\textbf{$C_3$-gold} & \textbf{20} & \textbf{.516} & \textbf{+7} & $\boldsymbol{0.016}$ \\
\bottomrule
\end{tabular}
\end{table}

All paired exact tests are computed at the family level. The slice is small, so we treat the held-out result as a directional replication rather than a fully-powered confirmation.

\section{Replication on MATH500 and AIME 2024/25}
\label{app:second_datasets}

This appendix runs the full OracleLadder protocol on two public datasets outside NuminaMath: MATH500~\citep{hendrycksmath2021,lightman2023verify} and the AIME 2024 and 2025 competitions. All runs use one student, Qwen3-8B with thinking disabled~\citep{yang2025qwen3}. The same model is the screening anchor, the model whose direct failures decide which problems enter the set.

\paragraph{What stays the same.}
The teacher is GPT-5.4~\citep{openai2026gpt54}, the model that compiled the 354-family set. The six parent conditions use the prompt templates of Appendix~\ref{app:prompts}: $C_1$, $C_2$-correct, $C_2$-random, $C_2$-generic, $C_3$-gold, and $C_3$-mismatched. $C_2$-random and $C_3$-mismatched draw their substitute content from other families of the same dataset. The same symbolic verifier grades every parent and milestone answer. Each parent condition gets $K{=}8$ attempts at temperature 1.0 and \texttt{max\_tokens=16384}. The milestone test (Stage~0) gives each tested milestone $K{=}8$ attempts under the same settings. A family counts as recovered under a condition when at least one of its 8 attempts is correct (pass@8).

\paragraph{What changes.}
Four parts of the construction differ from the 354-family set. First, the parent problems come from MATH500 and AIME. Second, the student and anchor is Qwen3-8B (thinking disabled), served on Tinker, in place of Qwen-base. Third, the screen keeps a problem when Qwen3-8B fails all 4 direct attempts at \texttt{max\_tokens=16384}. The 354-family screen used 8 direct attempts and also required the anchor to pass the milestone test, so here the milestone test is purely a measurement. Fourth, the teacher compiles each family from the problem statement and its final answer, without a reference solution.

\paragraph{Structural filter.}
A family enters the analysis when it has at least two tested non-\textsc{Integrate} milestones, as in the 354-family set. Before counting, a leak rule drops every milestone whose gold answer the verifier accepts as the parent answer. The milestone test and the $C_2$ and $C_3$ prompts use only the remaining milestones. This leak rule removes one MATH500 family, whose second milestone has the parent answer $\frac{35}{64}$ as its gold answer. Among the kept families, it removes 14 MATH500 milestones and 6 AIME 2024/25 milestones.

\paragraph{Screening yields and families.}
Table~\ref{tab:second_construction} gives the construction counts. The screen keeps 36 of 500 MATH500 problems (7.2\%) and 38 of 60 AIME 2024/25 problems (63.3\%). The 38 AIME 2024/25 families split into 17 from 2024 and 21 from 2025. The AIME 2025 slice is a separate run on a second transcription of the same 30 AIME 2025 problems. The screen keeps 21 of these 30 (70.0\%). Nineteen of the 21 slice families come from problems that are also among the 21 AIME 2025 families of the AIME 2024/25 run. The slice therefore repeats the whole pipeline on largely the same problems, with a fresh screen, fresh roadmaps, and fresh attempts. The teacher returned a valid milestone family for all 95 kept problems. The structural filter then leaves 35 MATH500, 38 AIME 2024/25, and 21 AIME 2025 families.

\begin{table}[h]
\centering
\footnotesize
\caption{Construction counts and five-way taxonomy on the second datasets. The student and screening anchor is Qwen3-8B (thinking disabled). Tested milestones are the non-\textsc{Integrate} milestones left after the leak rule. The six taxonomy rows count families and sum to the family count of each column.}
\label{tab:second_construction}
\vspace{4pt}
\begin{tabular}{lrrr}
\toprule
 & \textbf{MATH500} & \textbf{AIME 2024/25} & \textbf{AIME 2025 slice} \\
\midrule
Problems screened & 500 & 60 & 30 \\
Kept by the screen (0 of 4 correct) & 36 (7.2\%) & 38 (63.3\%) & 21 (70.0\%) \\
Valid milestone families & 36 & 38 & 21 \\
Families after the structural filter & 35 & 38 & 21 \\
Mean tested milestones per family & 4.7 & 5.2 & 6.1 \\
Tested milestones solved alone ($\geq$1 of 8) & 47/163 & 75/199 & 43/128 \\
Families passing the milestone test & 0 & 1 & 2 \\
\midrule
\textsc{Direct} & 16 & 2 & 5 \\
\textsc{Roadmap-gap} & 0 & 0 & 0 \\
\textsc{Milestone-execution-gap} & 0 & 0 & 0 \\
\textsc{Composition-gap} & 0 & 1 & 1 \\
\textsc{Missing-milestone-gap} & 6 & 16 & 3 \\
\textsc{Capability-gap} & 13 & 19 & 12 \\
\midrule
\textsc{Unrecovered} (composition $+$ capability) & 13 (37\%) & 20 (53\%) & 13 (62\%) \\
\bottomrule
\end{tabular}
\end{table}

\paragraph{The matched roadmap drives recovery.}
Table~\ref{tab:second_recovery} reports family-level recovery and per-attempt success for Qwen3-8B (thinking disabled) under all six conditions. On AIME 2024/25, $C_2$-correct recovers 17 of 38 families (45\%). $C_2$-random and $C_2$-generic each recover 5 of 38 (13\%), and $C_1$ recovers 2 of 38 (5\%). Paired exact McNemar tests at the family level give $p{=}0.004$ for $C_2$-correct against $C_2$-random and $p{=}0.002$ against $C_2$-generic. Per attempt, $C_2$-correct succeeds on 74 of 304 attempts (24.3\%), against 7 of 304 (2.3\%) for each control.

\begin{table}[h]
\centering
\footnotesize
\caption{Recovery under the six conditions for Qwen3-8B (thinking disabled). \emph{Fam.} counts families with at least 1 correct attempt out of 8 (pass@8). \emph{Attempts} gives the correct attempts and their share, out of 280 (MATH500), 304 (AIME 2024/25), and 168 (AIME 2025 slice) attempts per condition.}
\label{tab:second_recovery}
\vspace{4pt}
\setlength{\tabcolsep}{4pt}
\begin{tabular}{lrrrrrr}
\toprule
& \multicolumn{2}{c}{\textbf{MATH500} ($n{=}35$)} & \multicolumn{2}{c}{\textbf{AIME 2024/25} ($n{=}38$)} & \multicolumn{2}{c}{\textbf{AIME 2025 slice} ($n{=}21$)} \\
\cmidrule(lr){2-3} \cmidrule(lr){4-5} \cmidrule(lr){6-7}
\textbf{Condition} & Fam. & Attempts & Fam. & Attempts & Fam. & Attempts \\
\midrule
$C_1$ direct & 16 & 28 (10.0\%) & 2 & 3 (1.0\%) & 5 & 6 (3.6\%) \\
$C_2$-random & 12 & 22 (7.9\%) & 5 & 7 (2.3\%) & 2 & 3 (1.8\%) \\
$C_2$-generic & 16 & 27 (9.6\%) & 5 & 7 (2.3\%) & 4 & 8 (4.8\%) \\
\textbf{$C_2$-correct} & \textbf{18} & \textbf{106 (37.9\%)} & \textbf{17} & \textbf{74 (24.3\%)} & \textbf{6} & \textbf{24 (14.3\%)} \\
$C_3$-mismatched & 17 & 91 (32.5\%) & 14 & 65 (21.4\%) & 5 & 16 (9.5\%) \\
\textbf{$C_3$-gold} & \textbf{16} & \textbf{107 (38.2\%)} & \textbf{15} & \textbf{84 (27.6\%)} & \textbf{6} & \textbf{34 (20.2\%)} \\
\bottomrule
\end{tabular}
\end{table}

On MATH500, per-attempt success separates the conditions most clearly. $C_2$-correct succeeds on 106 of 280 attempts (37.9\%), 3.8 times the $C_1$ rate of 28 of 280 (10.0\%). $C_2$-random and $C_2$-generic stay at 22 of 280 (7.9\%) and 27 of 280 (9.6\%). The MATH500 pass@8 counts sit close together because of how the families were selected. Every MATH500 family failed 4 screening attempts, yet $C_1$ recovers 16 of 35 families at pass@8. A family with a per-attempt success rate near 10\% often fails 4 attempts and then succeeds once in 8 new ones.

Table~\ref{tab:second_threshold} therefore repeats the threshold analysis of Appendix~\ref{app:threshold}. A family counts as recovered when at least $\tau$ of its 8 attempts are correct. At $\tau{=}3$, $C_2$-correct recovers 15 of 35 MATH500 families, against 2 for each control and 3 for $C_1$. Paired exact McNemar tests give $p{<}0.001$ for $C_2$-correct against each of these three conditions. On AIME 2024/25, $C_2$-correct recovers more families than $C_1$, $C_2$-random, and $C_2$-generic at every $\tau$ from 1 to 4.

\begin{table}[h]
\centering
\footnotesize
\caption{Family-level recovery as the success threshold $\tau$ varies, for Qwen3-8B (thinking disabled). A family counts as recovered under a condition when at least $\tau$ of its 8 attempts are correct. The column $\tau{=}1$ is pass@8.}
\label{tab:second_threshold}
\vspace{4pt}
\setlength{\tabcolsep}{5pt}
\begin{tabular}{lrrrrrrrr}
\toprule
& \multicolumn{4}{c}{\textbf{MATH500} ($n{=}35$)} & \multicolumn{4}{c}{\textbf{AIME 2024/25} ($n{=}38$)} \\
\cmidrule(lr){2-5} \cmidrule(lr){6-9}
\textbf{Condition} & $\tau{=}1$ & $\tau{=}2$ & $\tau{=}3$ & $\tau{=}4$ & $\tau{=}1$ & $\tau{=}2$ & $\tau{=}3$ & $\tau{=}4$ \\
\midrule
$C_1$ direct & 16 & 8 & 3 & 1 & 2 & 1 & 0 & 0 \\
$C_2$-random & 12 & 5 & 2 & 2 & 5 & 1 & 1 & 0 \\
$C_2$-generic & 16 & 7 & 2 & 1 & 5 & 1 & 1 & 0 \\
\textbf{$C_2$-correct} & \textbf{18} & \textbf{16} & \textbf{15} & \textbf{13} & \textbf{17} & \textbf{13} & \textbf{11} & \textbf{10} \\
$C_3$-mismatched & 17 & 15 & 13 & 12 & 14 & 12 & 9 & 9 \\
\textbf{$C_3$-gold} & \textbf{16} & \textbf{15} & \textbf{15} & \textbf{14} & \textbf{15} & \textbf{13} & \textbf{11} & \textbf{10} \\
\bottomrule
\end{tabular}
\end{table}

The AIME 2025 slice points the same way at a smaller size. $C_2$-correct recovers 6 of 21 families, against 2 of 21 for $C_2$-random, 4 of 21 for $C_2$-generic, and 5 of 21 for $C_1$. With 21 families, the paired exact McNemar test against $C_2$-random gives $p{=}0.125$. Per attempt, $C_2$-correct succeeds on 24 of 168 attempts (14.3\%), against 6 of 168 (3.6\%) for $C_1$.

Gold milestone answers add little on top of the matched roadmap on these datasets. On MATH500, $C_3$-gold succeeds on 107 of 280 attempts and $C_2$-correct on 106 of 280. On AIME 2024/25, the counts are 84 and 74 of 304. $C_3$-mismatched stays close behind $C_3$-gold, at 91 of 280 and 65 of 304. On the 354-family set, $C_3$-gold solves more families than $C_2$-correct for four of the six panel models (Table~\ref{tab:panel_details}). The gap between $C_2$-correct and $C_3$-gold is therefore itself a per-dataset reading.

\paragraph{Five-way taxonomy.}
Table~\ref{tab:second_construction} also crosses the milestone test with the parent-probe outcome. Qwen3-8B (thinking disabled) passes the milestone test on 0 of 35 MATH500 families, 1 of 38 AIME 2024/25 families, and 2 of 21 AIME 2025 families. The \textsc{composition-gap} therefore holds 0, 1, and 1 families. Unrecovered families that fail the milestone test fall into \textsc{capability-gap} (13, 19, and 12 families). Families that $C_2$ or $C_3$ rescues after a failed milestone test fall into \textsc{missing-milestone-gap} (6, 16, and 3 families).

For Qwen3-8B (thinking disabled), no probe solves 13 of 35 MATH500 families (37\%), 20 of 38 AIME 2024/25 families (53\%), and 13 of 21 AIME 2025 families (62\%). On the 354-family set, this unrecovered share is 40--64\% across the six panel models (Table~\ref{tab:stage0}). The two sets differ in where these families sit. On the 354-family set, most of them pass the milestone test and land in \textsc{composition-gap}. Here, almost all of them fail it and land in \textsc{capability-gap}.

Two measured differences lie behind the low pass counts. The teacher writes more milestones per family here, with 4.7, 5.2, and 6.1 tested milestones against 2.2 on the 354-family set. Qwen3-8B (thinking disabled) also solves fewer milestones alone: 47 of 163 on MATH500 (29\%), 75 of 199 on AIME 2024/25 (38\%), and 43 of 128 on AIME 2025 (34\%). The milestone test requires every tested milestone, so the two effects compound.

Gold milestone answers are also longer here. Their median length is 88 characters on MATH500, 143 on AIME 2024/25, and 129 on the AIME 2025 slice, against 13 on the 354-family set. Many state a result together with a short justification. Milestone solve rates are similar for short and long gold answers, pooled over the 79 families with no leak-filtered milestone. Qwen3-8B (thinking disabled) solves 16 of 48 milestones (33\%) with gold answers of at most 20 characters and 110 of 313 (35\%) above 60 characters.

Category shares therefore depend on the dataset and the student. The matched-versus-corrupted roadmap contrast is the quantity that replicates across datasets. For this reason, we report \textsc{composition-gap} counts together with the student's milestone-test pass count.

\paragraph{Anchor dependence of the screen.}
Table~\ref{tab:second_anchor_yield} applies the same screen with two more anchors, gpt-oss-20b~\citep{openai2025gptoss} and Llama-3.2-1B-Instruct. The rule and prompts stay fixed: a problem is kept when the anchor fails all 4 direct attempts. On MATH500, the share kept ranges from 5 of 500 (1.0\%) under gpt-oss-20b to 345 of 500 (69.0\%) under Llama-3.2-1B-Instruct. On AIME 2024/25, it ranges from 10 of 60 (16.7\%) to 58 of 60 (96.7\%). The anchor and the dataset jointly define the set that a screen selects.

\begin{table}[h]
\centering
\footnotesize
\caption{The same screen under three anchors. A problem is kept when the anchor fails all 4 direct attempts at \texttt{max\_tokens=16384}. \emph{Probed families kept} counts the families of Table~\ref{tab:second_recovery} that this anchor's screen also keeps. Qwen3-8B keeps all of them by construction. Qwen3-8B (thinking disabled) and gpt-oss-20b run on Tinker. Llama-3.2-1B-Instruct runs on a hosted inference API with the provider's chat template. For gpt-oss-20b on MATH500, a restarted run logged some attempts twice, and we keep the first record of each attempt.}
\label{tab:second_anchor_yield}
\vspace{4pt}
\begin{tabular}{lrrrr}
\toprule
& \multicolumn{2}{c}{\textbf{Problems kept}} & \multicolumn{2}{c}{\textbf{Probed families kept}} \\
\cmidrule(lr){2-3} \cmidrule(lr){4-5}
\textbf{Anchor} & MATH500 & AIME 2024/25 & MATH500 & AIME 2024/25 \\
\midrule
gpt-oss-20b & 5/500 (1.0\%) & 10/60 (16.7\%) & 4/35 & 9/38 \\
Qwen3-8B (thinking disabled) & 36/500 (7.2\%) & 38/60 (63.3\%) & 35/35 & 38/38 \\
Llama-3.2-1B-Instruct & 345/500 (69.0\%) & 58/60 (96.7\%) & 33/35 & 38/38 \\
\bottomrule
\end{tabular}
\end{table}

The Llama-3.2-1B-Instruct failure set contains all 38 AIME 2024/25 families and 33 of the 35 MATH500 families. The two MATH500 families outside it have 0 of 8 correct attempts under all six conditions. Every family count in Tables~\ref{tab:second_recovery} and~\ref{tab:second_threshold} is therefore unchanged on the contained families. The recovery contrast thus holds on families that a non-Qwen screen also keeps. The gpt-oss-20b failure set contains 4 MATH500 and 9 AIME 2024/25 families, which is too few for a contrast.

A 30-problem MATH500 pilot with 2 attempts per problem ruled out two other candidate anchors. Llama-3.2-3B put its answer in \texttt{\textbackslash boxed\{\}} in 28 of 60 attempts and solved 0 of 60. Nemotron-3-Nano-30B-A3B solved 58 of 60 attempts, which would leave almost no problems to screen in.

\section{Code-Generation Pilot on LiveCodeBench}
\label{app:code_pilot}

This appendix instantiates OracleLadder for code generation on LiveCodeBench~\citep{jain2025livecodebench}. Running the program on test cases replaces the symbolic verifier, and helper functions replace math milestones. The pilot covers 10 problems and one student, Qwen3-8B with thinking disabled. Table~\ref{tab:code_instantiation} maps each protocol component to its code version.

\begin{table}[h]
\centering
\small
\caption{OracleLadder components in math and their counterparts in the LiveCodeBench pilot.}
\label{tab:code_instantiation}
\vspace{4pt}
\begin{tabular}{@{}p{0.2\linewidth}p{0.34\linewidth}p{0.4\linewidth}@{}}
\toprule
\textbf{Component} & \textbf{Math} & \textbf{Code} \tabularnewline
\midrule
\raggedright Parent problem & \raggedright Competition problem with a final answer & \raggedright LiveCodeBench problem that reads standard input, with test cases \tabularnewline
\raggedright Verifier & \raggedright Symbolic check of the boxed answer against the gold answer & \raggedright Run the program on up to 12 test cases and compare its output with the expected output \tabularnewline
\raggedright Milestone & \raggedright Sub-goal with a checkable gold answer & \raggedright Helper function with a signature, a docstring, at least 2 assert tests, and a gold implementation \tabularnewline
\raggedright Milestone test (Stage~0) & \raggedright Solve each milestone alone & \raggedright Implement each helper from its signature, docstring, and tests, graded by its asserts \tabularnewline
\raggedright $C_1$ & \raggedright Parent problem only & \raggedright Problem statement only \tabularnewline
\raggedright $C_2$ & \raggedright Parent plus the roadmap of milestone descriptions & \raggedright Statement plus helper signatures, docstrings, and an overall plan \tabularnewline
\raggedright $C_3$ & \raggedright Roadmap plus gold milestone answers & \raggedright Statement, overall plan, and the verified gold helper implementations \tabularnewline
\raggedright Teacher & \raggedright GPT-5.4 Thinking & \raggedright Inkling \tabularnewline
\raggedright Compile-time check & \raggedright Schema and leak checks on each milestone & \raggedright Every gold helper passes its asserts, and the full gold program passes the parent tests \tabularnewline
\bottomrule
\end{tabular}
\end{table}

\paragraph{Construction.}
The pool is the 78 medium-difficulty LiveCodeBench problems that read standard input and have at least two test cases. Qwen3-8B (thinking disabled) fails all 4 direct attempts on 31 of them. For these 31 problems, the Inkling model from Thinking Machines compiles a helper-function roadmap. Each packet holds helper functions, an overall plan, and a full gold program. A packet is accepted when every gold helper passes its own asserts and the gold program passes the parent tests. A rejected packet returns to the teacher with the error message, for up to 3 attempts. Ten of the 31 problems yield an accepted packet, with 39 helpers in total (3 or 4 per problem).

\paragraph{Probing.}
Each condition gets $K{=}8$ attempts from Qwen3-8B (thinking disabled) at temperature 0.7 and \texttt{max\_tokens=16384}. A program passes when its output matches the expected output on every executed test case, with up to 12 cases per problem. The helper test, the code version of the milestone test, gives each helper 8 attempts and grades each attempt by that helper's asserts. The pilot ran $C_1$, $C_2$, $C_3$, and the helper test. $C_1$ completed 62 of its 80 planned attempts, and 5 of the 10 problems have all 8.

\begin{table}[h]
\centering
\small
\caption{Code-pilot results for Qwen3-8B (thinking disabled) on 10 LiveCodeBench problems, graded by test execution. \emph{Problems recovered} counts problems with at least 1 passing attempt out of 8. $C_1$ completed 62 of 80 planned attempts, so its rate uses 62 attempts.}
\label{tab:code_results}
\vspace{4pt}
\begin{tabular}{lrrr}
\toprule
\textbf{Condition} & \textbf{Passing attempts} & \textbf{Rate} & \textbf{Problems recovered} \\
\midrule
$C_1$ direct & 4/62 & 6.5\% & 1/10 \\
$C_2$ helper plan & 33/80 & 41.2\% & 6/10 \\
$C_3$ plan $+$ verified gold helpers & 71/80 & 88.8\% & 10/10 \\
\bottomrule
\end{tabular}
\end{table}

\paragraph{Results.}
Table~\ref{tab:code_results} shows the ladder of help in the code domain. Per-attempt success rises from 4 of 62 (6.5\%) with no help to 33 of 80 (41.2\%) with the helper plan. It reaches 71 of 80 (88.8\%) with the verified gold helpers. Problem-level recovery rises from 1 to 6 to 10 of 10. The one $C_1$ success, \texttt{abc318\_d}, failed all 4 screening attempts and then passed 4 of 8 probe attempts. This is the same screen-and-probe sampling effect seen on MATH500 (Appendix~\ref{app:second_datasets}). One problem, \texttt{abc335\_c}, passes more often with the plan alone (7 of 8) than with the gold helpers (2 of 8).

In the helper test, Qwen3-8B (thinking disabled) passes 208 of 312 attempts (66.7\%) and solves 31 of 39 helpers at least once. Four of the 10 problems pass the helper test. Crossing it with the three conditions gives 1 \textsc{Direct}, 3 \textsc{roadmap-gap}, 1 \textsc{milestone-execution-gap}, and 5 \textsc{missing-milestone-gap} problems. $C_3$ recovers all 10 problems, so \textsc{composition-gap} and \textsc{capability-gap} are both empty.

\paragraph{What the pilot shows.}
The protocol carries over to a second domain with a different deterministic verifier. Each math component has a direct code counterpart. The code roadmap is also checked by construction, because the teacher's gold program must pass the tests before the packet is accepted. The three help levels order the same way as in math, with large gains from the roadmap and from verified intermediate results. The specificity contrast against $C_2$-random, $C_2$-generic, and $C_3$-mismatched comes from the math datasets (Section~\ref{sec:oracle_specificity} and Appendix~\ref{app:second_datasets}). The code pilot measures the help ladder on 10 problems with one student.

\section{Teacher and Granularity Robustness}
\label{app:teacher_robust}

This appendix measures how much the readout depends on who writes the roadmaps and how finely they are written. We change the roadmaps in two ways, with a second teacher and with coarser or finer roadmaps from the same teacher, and keep the student fixed at gpt-oss-20b. Per-family $C_2$ and $C_3$ recovery stays the same for 83--87\% of families under the second teacher and for 85\% under a roadmap four times finer. Five-way labels agree less often, because they also depend on the milestone test (Stage~0), which is measured on one specific roadmap.

\paragraph{Claude Sonnet 4.6 pilot.}
A first pilot re-compiled 40 parent problems with Claude Sonnet 4.6 as the teacher and kept the rest of the pipeline fixed. After the construction screen and the two-milestone filter, 8 milestone families remained. On these 8, MilestoneRL-2K succeeds on 42.2\% of $C_2$-correct attempts and 46.9\% of $C_3$-gold attempts, against 9.4\% under $C_1$, 14.1\% under $C_2$-random, and 6.2\% under $C_2$-generic, so the specificity pattern appears even in this small pilot.

\subsection{A Second Teacher}

\paragraph{Setup.}
The second teacher is Inkling, an open-weights model from Thinking Machines with 975B total and 41B active parameters~\citep{tml2026inkling}, served through Tinker. We sampled 60 of the 354 milestone families at random (seed 42) and ran the same compilation code in its decomposition mode at temperature 0.1. Inkling saw each parent problem and its reference answer. The GPT-5.4 compilation of the released set also saw the reference solution $G(P)$, so the two teachers differ in their input as well as in their weights. Inkling returned a valid roadmap for 53 of the 60 families, and its output for the other 7 was not valid JSON. As in the main protocol, the \textsc{Integrate} step stays private, and a leak rule removes every milestone whose gold answer the verifier accepts as the reference answer. The $C_2$-correct and $C_3$-gold prompts use the remaining milestones.

\paragraph{Structure.}
Table~\ref{tab:teacher_structure} compares the two teachers' roadmaps on the 53 families. Inkling writes longer roadmaps than GPT-5.4, 8.1 milestones per family against 5.8, with a similar mix of milestone types. The GPT-5.4 column counts its full decomposition from the decomposition-mode run that recovered the \textsc{Integrate} steps (Appendix~\ref{app:dataset}), because the released families keep only the tested milestones. The roadmaps shown to the student differ more, with 6.8 tested milestones per family from Inkling against 2.1 in the released families. Step-by-step alignment between two roadmaps is ill-defined when one coarse milestone spans several fine ones, so we compare length and type mix.

\begin{table}[h]
\centering
\footnotesize
\caption{Roadmap structure under the two teachers on the 53 families with a valid Inkling roadmap. Type shares count all milestones, including the private \textsc{Integrate} step. Tested milestones are the non-\textsc{Integrate} milestones left after the leak rule, which the $C_2$-correct and $C_3$-gold prompts show.}
\label{tab:teacher_structure}
\vspace{4pt}
\begin{tabular}{lrr}
\toprule
 & \textbf{Inkling} & \textbf{GPT-5.4} \\
\midrule
Milestones per family, all types & 8.1 & 5.8 \\
Tested milestones per family & 6.8 & 2.1 \\
\midrule
\textsc{Compute} share & 23.0\% & 21.1\% \\
\textsc{Lemma} share & 14.3\% & 12.7\% \\
\textsc{Sanity} share & 12.9\% & 16.6\% \\
\textsc{Model} share & 12.6\% & 17.2\% \\
\textsc{Integrate} share & 12.4\% & 17.5\% \\
\textsc{Key\_Move} share & 10.8\% & 12.7\% \\
\textsc{Normalize} share & 8.9\% & 0.3\% \\
\textsc{Warmup} share & 5.2\% & 1.9\% \\
\bottomrule
\end{tabular}
\end{table}

\paragraph{Per-family recovery.}
Table~\ref{tab:roadmap_agreement} gives the behavioral comparison. With $K{=}8$ attempts at \texttt{max\_tokens=16384}, gpt-oss-20b reaches the same $C_2$ outcome under both teachers' roadmaps for 44 of 53 families (83\%), the same $C_3$ outcome for 46 (87\%), and the same pair of outcomes for 41 (77\%). Most disagreements sit at the success threshold. In 5 of the 9 $C_2$ disagreements and 5 of the 7 $C_3$ disagreements, one roadmap gives 1 correct attempt out of 8 and the other gives none. The level of recovery shifts slightly. Inkling's roadmaps recover 20 families under $C_2$ and 19 under $C_3$, against 23 and 24 under the released roadmaps, so recovery levels are best compared under one teacher.

\begin{table}[h]
\centering
\footnotesize
\caption{Per-family recovery by gpt-oss-20b under two roadmaps of the same families ($K{=}8$, \texttt{max\_tokens=16384}). A family is recovered under a condition when at least 1 of its 8 attempts is correct. Left: released GPT-5.4 roadmaps against Inkling roadmaps. Right: coarse against fine GPT-5.4 roadmaps from the controlled ablation. Pairs of counts list the first roadmap, then the second.}
\label{tab:roadmap_agreement}
\vspace{4pt}
\begin{tabular}{lcc}
\toprule
 & \textbf{Released vs.\ Inkling} & \textbf{Coarse vs.\ fine} \\
 & (53 families) & (33 families) \\
\midrule
Tested milestones per family & 2.1 / 6.8 & 2.0 / 8.1 \\
Families recovered under $C_2$-correct & 23 / 20 & 11 / 10 \\
Families recovered under $C_3$-gold & 24 / 19 & 13 / 12 \\
Same $C_2$ outcome & 44 (83\%) & 28 (85\%) \\
Same $C_3$ outcome & 46 (87\%) & 28 (85\%) \\
Same $C_2$ and $C_3$ outcomes & 41 (77\%) & 25 (76\%) \\
Disagreements at 1 vs.\ 0 of 8 ($C_2$, $C_3$) & 5 of 9, 5 of 7 & 2 of 5, 3 of 5 \\
\bottomrule
\end{tabular}
\end{table}

\subsection{Five-Way Labels Under the Second Teacher}

\paragraph{Gradable roadmaps.}
Five-way labels need the milestone test, and the milestone test needs gold milestone answers that the verifier can grade. Inkling's gold answers are often compound statements. Only 22\% of them pass a simple atomicity check, which asks for one value, expression, equation, interval, or tuple without prose, against 98\% of the gold answers in the released families, and their median length is 48 characters against 13. We therefore added an answer-format instruction with good and bad examples to Inkling's teacher prompt, leaving the released prompt unchanged. We re-compiled 45 of the 50 families that have released unaided attempts (Appendix~\ref{app:trace}). Inkling returned valid roadmaps for 39, with 79\% of their gold answers atomic and a median length of 19 characters. On these 39 families we ran the milestone test and $C_1$, $C_2$-correct, and $C_3$-gold for gpt-oss-20b, all with $K{=}8$ attempts.

\paragraph{Agreement.}
The five-way labels agree for 22 of the 39 families under the strict milestone test. They agree for 29 of 39 when the milestone test is scored by the fraction of milestones passed, with a family counted as passing when the student solves at least half of its tested milestones alone (Table~\ref{tab:fiveway_transitions}).

\begin{table}[h]
\centering
\footnotesize
\caption{Five-way labels of gpt-oss-20b on 39 families under the released GPT-5.4 roadmaps and under Inkling roadmaps with atomic gold answers. Rows list each transition from the released label to the Inkling label. The strict milestone test requires every tested milestone. The fraction score requires at least half of them. $C_1$ is sampled afresh under the Inkling roadmaps.}
\label{tab:fiveway_transitions}
\vspace{4pt}
\begin{tabular}{llcc}
\toprule
\textbf{Released label} & \textbf{Inkling label} & \textbf{Strict test} & \textbf{Fraction score} \\
\midrule
\textsc{Direct} & \textsc{Direct} & 22 & 22 \\
\textsc{Composition-gap} & \textsc{Composition-gap} & 0 & 7 \\
\textsc{Composition-gap} & \textsc{Capability-gap} & 11 & 4 \\
\textsc{Composition-gap} & \textsc{Missing-milestone-gap} & 1 & 0 \\
\textsc{Composition-gap} & \textsc{Roadmap-gap} & 0 & 1 \\
\textsc{Composition-gap} & \textsc{Direct} & 0 & 1 \\
\textsc{Roadmap-gap} & \textsc{Direct} & 2 & 2 \\
\textsc{Roadmap-gap} & \textsc{Composition-gap} & 1 & 2 \\
\textsc{Roadmap-gap} & \textsc{Capability-gap} & 1 & 0 \\
\textsc{Capability-gap} & \textsc{Direct} & 1 & 0 \\
\midrule
\multicolumn{2}{l}{Families with the same label} & 22 & 29 \\
\multicolumn{2}{l}{Families passing the milestone test (released / Inkling)} & 35 / 2 & 38 / 25 \\
\bottomrule
\end{tabular}
\end{table}

\paragraph{Why the labels move.}
The disagreements come from the milestone test. Under the strict test, 11 of the 17 disagreements are families that stay \textsc{Unrecovered} under both roadmaps, so their $C_1$, $C_2$, and $C_3$ outcomes agree, and only the milestone test flips, from pass (\textsc{composition-gap}) to fail (\textsc{capability-gap}). The milestone test requires every tested milestone, and the two roadmaps make that requirement very different. The Inkling roadmaps have 7.0 tested milestones per family against 2.2, so each family has more chances to fail. gpt-oss-20b also solves a smaller share of them alone, 145 of 274 (53\%) against 82 of 87 (94\%) of the released milestones, which passed the construction screen where Qwen-base had to solve every one of them. Together these effects let the strict test pass for 2 of the 39 families under the Inkling roadmaps and for 35 of 39 under the released roadmaps. The fraction score depends less on roadmap length. It passes 25 and 38 families and raises the agreement to 29 of 39.

\subsection{Controlled Granularity Ablation}

\paragraph{Setup.}
To separate granularity from the teacher, we kept GPT-5.4 as the teacher and changed only the requested number of milestones. For 40 families sampled at random (seed 42), one added sentence asked for exactly 3 milestones with large steps (coarse) or for 8 to 10 milestones with small steps (fine), and everything else stayed the same. As for Inkling, the teacher saw the parent problem and its reference answer. The teacher followed the request, returning 2 non-\textsc{Integrate} milestones for every coarse roadmap and 8.2 per family (range 7 to 9) for the fine roadmaps. The leak rule leaves at least two tested milestones under both arms for 33 families, because 7 coarse roadmaps have a milestone whose answer equals the reference answer. The released roadmaps average 2.2 tested milestones, so the coarse arm is close to a re-compilation at the released granularity and the fine arm is about four times finer.

\paragraph{Result.}
Per-family recovery by gpt-oss-20b is as stable under the granularity change as under the teacher change (Table~\ref{tab:roadmap_agreement}). The coarse and fine roadmaps give the same $C_2$ outcome for 28 of 33 families (85\%), the same $C_3$ outcome for 28 (85\%), and the same pair of outcomes for 25 (76\%). Aggregate recovery moves by one family, 11 against 10 under $C_2$ and 13 against 12 under $C_3$. Per-attempt success is 18.2\% under $C_2$ and 17.0\% under $C_3$ for the coarse roadmaps, against 17.4\% and 20.1\% for the fine roadmaps.

\subsection{How to Compare Readouts Across Roadmaps}

The two experiments split the readout into a part that transfers across roadmaps and a part that belongs to one roadmap. Per-family $C_2$ and $C_3$ recovery agrees for 83--87\% of families under a new teacher and for 85\% under a roadmap four times finer, so recovery is the quantity to compare across teachers. The five-way label also uses the milestone test, which requires every milestone of the roadmap it is run on, so a finer roadmap makes the test stricter. We therefore compare five-way labels only between roadmaps of matched granularity, and every five-way count in this paper uses the released roadmaps. Every milestone-test result is reported together with the number of tested milestones per family and the per-milestone solve rate, and the fraction-of-milestones score serves as a secondary readout when roadmap lengths differ.

\section{Milestone Answers in the Models' Own Solutions}
\label{app:trace}

This appendix asks whether the milestones of a roadmap are steps that models take when they solve the parent problem without help. Milestone answers appear in the models' unaided solutions 2.9 to 3.9 times as often as in solutions to other problems, and 6.6 to 12.6 times as often when we count only answers that the problem statement does not already contain. The analysis reads the released unaided ($C_1$) attempts of the format audit (Appendix~\ref{app:decoding_budget}), which cover 50 families drawn at random (seed 42) with 8 attempts each for Qwen-base, MilestoneRL-2K, Llama-3.3-70B-Instruct, and DeepSeek-V3.1 at \texttt{max\_tokens=4096}.

\paragraph{Containment test.}
A milestone counts as attained in an attempt when its gold answer appears in the attempt's text, verbatim, after the verifier's string normalization, or as a matching number. The test is deterministic. It is also conservative, because an attempt can reach a milestone in another form, so the rates below are lower bounds on how often the attempts pass through the milestones. The cross-problem control checks the same gold answers against the attempts on every other family of the 50. It keeps the gold values fixed and so measures how often a value appears in an unrelated math solution. Six of the 110 gold answers (5.5\%) already appear in their problem statement, and an attempt that restates the problem contains them. We therefore also report the rates over the 104 statement-novel gold answers.

\paragraph{Milestone answers appear in the models' own solutions.}
Table~\ref{tab:trace_containment} gives the rates. Over all gold answers, 13.6--17.2\% of milestone and attempt pairs contain the milestone's answer, against 3.8--5.2\% for the cross-problem control. Over statement-novel gold answers the rate is 9.4--12.5\% against 0.8--1.7\%.

\begin{table}[h]
\centering
\footnotesize
\caption{Share of milestone and attempt pairs whose unaided ($C_1$) attempt contains the milestone's gold answer, on 50 families with 8 attempts each. The control checks the same gold answers against the attempts on the other 49 families. Statement-novel gold answers are the 104 of 110 that the problem statement does not contain.}
\label{tab:trace_containment}
\vspace{4pt}
\setlength{\tabcolsep}{4pt}
\begin{tabular}{lrrrrrr}
\toprule
 & \multicolumn{3}{c}{\textbf{All gold answers}} & \multicolumn{3}{c}{\textbf{Statement-novel gold answers}} \\
\cmidrule(lr){2-4} \cmidrule(lr){5-7}
\textbf{Model} & \textbf{Own} & \textbf{Control} & \textbf{Ratio} & \textbf{Own} & \textbf{Control} & \textbf{Ratio} \\
\midrule
Qwen-base & 13.6\% & 3.8\% & 3.6$\times$ & 9.4\% & 0.8\% & 12.3$\times$ \\
MilestoneRL-2K & 17.2\% & 4.4\% & 3.9$\times$ & 12.5\% & 1.2\% & 10.7$\times$ \\
Llama-3.3-70B-Instruct & 14.5\% & 4.3\% & 3.4$\times$ & 9.9\% & 0.8\% & 12.6$\times$ \\
DeepSeek-V3.1 & 15.2\% & 5.2\% & 2.9$\times$ & 11.3\% & 1.7\% & 6.6$\times$ \\
\bottomrule
\end{tabular}
\end{table}

\paragraph{Attainment tracks what each model solves alone.}
For three of the four models, a milestone that the model solves in its own milestone test appears in its unaided attempts far more often than a milestone it cannot solve alone (Table~\ref{tab:trace_stage0}). The rates are 14.6\% against 0.0\% for Qwen-base, 18.6\% against 1.4\% for MilestoneRL-2K, and 17.7\% against 2.8\% for DeepSeek-V3.1. Llama-3.3-70B-Instruct shows no such link, 14.3\% against 17.2\%, and only 8 of the 110 milestones are ones it cannot solve alone. Milestones that a model solves alone also tend to be easier, and easier results appear more often in any solution. The link therefore shows that the roadmap's difficulty profile follows each model's own, and it leaves open which steps a model computes internally.

\begin{table}[h]
\centering
\footnotesize
\caption{Share of milestone and attempt pairs whose unaided attempt contains the milestone's gold answer, split by whether the same model solves that milestone in its 16K milestone test ($\geq$1 of 8 attempts). Counts are attained pairs over all pairs.}
\label{tab:trace_stage0}
\vspace{4pt}
\begin{tabular}{lrr}
\toprule
\textbf{Model} & \textbf{Solved alone} & \textbf{Not solved alone} \\
\midrule
Qwen-base & 14.6\% (120/824) & 0.0\% (0/56) \\
MilestoneRL-2K & 18.6\% (150/808) & 1.4\% (1/72) \\
Llama-3.3-70B-Instruct & 14.3\% (117/816) & 17.2\% (11/64) \\
DeepSeek-V3.1 & 17.7\% (130/736) & 2.8\% (4/144) \\
\bottomrule
\end{tabular}
\end{table}

\paragraph{An independent roadmap fits the solutions as well.}
A model's unaided solution could follow this roadmap in particular, or any valid roadmap of the same problem could fit it equally well. We compare the released roadmaps with the Inkling roadmaps of the same 39 families (Appendix~\ref{app:teacher_robust}), whose gold answers were written in the same atomic format, on the same unaided attempts. The released gold answers are shorter, with a median of 15 characters against 19, and short strings are easier to find in a text. At raw length the released answers appear 1.02 to 1.51 times as often as Inkling's (Table~\ref{tab:trace_lengthmatched}). When both sets are restricted to gold answers of at most 25 characters, the ratio falls to 0.76--1.10, and at 15 characters or fewer it is 0.80--0.93.

\begin{table}[h]
\centering
\footnotesize
\caption{Containment rate of the released gold answers divided by the containment rate of Inkling's gold answers for the same 39 families, on the same unaided attempts. Each column restricts both sets of gold answers to the given length.}
\label{tab:trace_lengthmatched}
\vspace{4pt}
\begin{tabular}{lrrr}
\toprule
\textbf{Model} & \textbf{All lengths} & \textbf{$\leq$25 characters} & \textbf{$\leq$15 characters} \\
\midrule
Qwen-base & 1.28 & 0.92 & 0.82 \\
MilestoneRL-2K & 1.37 & 0.97 & 0.83 \\
Llama-3.3-70B-Instruct & 1.51 & 1.10 & 0.93 \\
DeepSeek-V3.1 & 1.02 & 0.76 & 0.80 \\
\bottomrule
\end{tabular}
\end{table}

\paragraph{Reading.}
The roadmap captures the structure of the problem, which is what makes cross-model comparison meaningful. Its milestones are quantities that models reach on their own when they work on this problem, far more often than on other problems, and an independent roadmap of the same problem fits the same solutions as well at matched answer length. The teacher compiles each family from the parent problem and its reference solution and never sees an evaluated model's output (Section~\ref{sec:method}). The same roadmap therefore serves as one fixed ruler for every model on the panel.

\section{Transition Matrix Counts}
\label{app:transition}

This appendix gives the family-level OutcomeRL-2K $\to$ MilestoneRL-2K parent-probe outcome transitions that back the longitudinal readout (Section~\ref{sec:results_training}). The endpoints disagree most on families that are not already direct solves under both arms, supporting the claim that the training arms reshape residual recovery structure rather than only shifting direct accuracy.

\begin{table}[h]
\centering
\small
\caption{OutcomeRL-2K $\to$ MilestoneRL-2K state transitions on the $n{=}354$ main set at \texttt{max\_tokens}${=}$16384, $K{=}8$.}
\label{tab:transition}
\vspace{4pt}
\setlength{\tabcolsep}{4pt}
\resizebox{\linewidth}{!}{%
\begin{tabular}{l|cccc|c}
\toprule
& \multicolumn{4}{c|}{\textbf{MilestoneRL-2K state}} & \\
\textbf{OutcomeRL-2K state} & \textsc{Direct} & \textsc{Roadmap-Needed} & \textsc{Answers-Needed} & \textsc{Unrecovered} & Row total \\
\midrule
\textsc{Direct}       & 84 & 13 & 5 & 4 & 106 \\
\textsc{Roadmap-Needed}    & 8 & 27 & 6 & 9 & 50 \\
\textsc{Answers-Needed}    & 1 & 3 & 6 & 4 & 14 \\
\textsc{Unrecovered}  & 3 & 4 & 5 & 172 & 184 \\
\midrule
Column total & 96 & 47 & 22 & 189 & 354 \\
\bottomrule
\end{tabular}}
\end{table}

\section{Audit Procedure Details}
\label{app:audit_details}

\begin{figure}[h]
\centering
\includegraphics[width=0.85\textwidth]{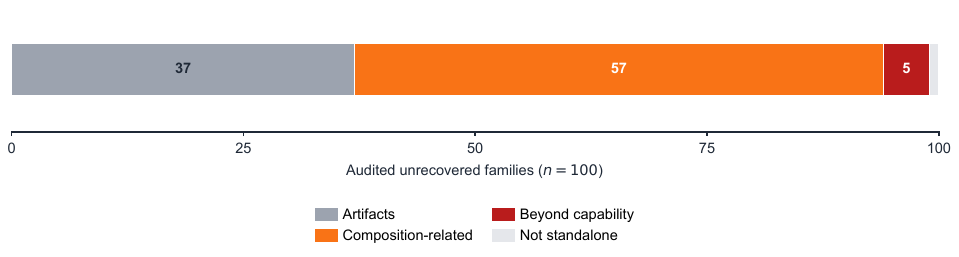}
\caption{\textbf{The unrecovered subset is mixed.}
Triage of the $n{=}100$ analyzed \textsc{Unrecovered} families under the leak-safe repair procedure. Bar segments show the mutually exclusive audit categories. The pessimistic upper bound on composition-related failures follows the bookkeeping in Appendix~\ref{app:audit_details}.}
\label{fig:audit_triage}
\end{figure}

This appendix gives the rubric, inter-rater statistics, and bookkeeping behind the author audit (Figure~\ref{fig:audit_triage}). Two raters independently classified 100 \textsc{Unrecovered} MilestoneRL-2K families with the rubric below. Disagreements were resolved by a fixed priority. The classified families were then run through two repair protocols (verifier-only and leak-safe decomposition), each followed by a fresh oracle re-evaluation. We report the rater rubric and the post-repair bookkeeping as two separate tables (Table~\ref{tab:audit_main}, Table~\ref{tab:audit_triage}), with mutually exclusive top-level categories in each.

\paragraph{Rubric (rater categories).} Each unrecovered family is classified into one of five mutually exclusive categories. (1)~\textsc{Genuine\_Composition}: milestones are well-formed, self-contained, and collectively sufficient. The failure is a real composition gap. (2)~\textsc{Decomposition\_Incomplete}: milestones are missing a critical composition or bridging step. Even with all milestone answers, the parent answer cannot be derived. (3)~\textsc{Verifier\_Artifact}: the parent gold answer is corrupted, malformed, or ambiguous. (4)~\textsc{Not\_Standalone}: one or more milestones reference undefined symbols from the parent. (5)~\textsc{Beyond\_Capability}: the problem is genuinely beyond the student model's capacity.

\paragraph{Disagreement resolution.} Priority rule (conservative, favoring data-quality explanations over composition-failure labels): \textsc{Verifier\_Artifact} $>$ \textsc{Beyond\_Capability} $>$ \textsc{Not\_Standalone} $>$ \textsc{Decomposition\_Incomplete} $>$ \textsc{Genuine\_Composition}.

\paragraph{Validation subset.} A separate human validation pass independently classifies a 50-family subset using the same rubric. Agreement with the primary labels, after the fixed priority rule, is $\kappa{=}0.64$~\citep{cohen1960kappa} (76\% raw agreement, ``substantial'' per \citet{landis1977interrater}). The validation pass finds slightly fewer genuine composition failures (12\% vs.\ 18\%) and more verifier artifacts (34\% vs.\ 26\%), consistent with a more conservative interpretation of milestone sufficiency.

\paragraph{Table A: rater rubric counts (raw human audit).} Table~\ref{tab:audit_main} reports the rubric category counts before any repair attempt. Categories are mutually exclusive and sum to 100.

\begin{table}[h]
\centering
\footnotesize
\caption{Human audit of 100 unrecovered families from the $n{=}354$ main set (inter-rater $\kappa{=}0.49$, validated at $\kappa{=}0.64$ on a 50-family subset). Categories are mutually exclusive.}
\label{tab:audit_main}
\vspace{4pt}
\setlength{\tabcolsep}{4pt}
\begin{tabular}{lrrll}
\toprule
\textbf{Rater category} & \textbf{Count} & \textbf{\%} & \textbf{Repair action} & \textbf{Leak-safe?} \\
\midrule
Decomposition incomplete & 50 & 50\% & Regenerate, leak-safe filter & \textbf{Yes} \\
Verifier artifact (bad canonical) & 26 & 26\% & Re-solve + replace canonical & \textbf{Yes} \\
\textbf{Genuine composition failure} & \textbf{18} & \textbf{18\%} & Keep as-is & --- \\
Beyond model capability & 5 & 5\% & Keep as-is & --- \\
Milestone not standalone & 1 & 1\% & Keep as-is & --- \\
\midrule
\textbf{Total} & \textbf{100} & 100\% & & \\
\bottomrule
\end{tabular}
\end{table}

\paragraph{Table B: post-repair triage and the upper bound (final interpretation).} Table~\ref{tab:audit_triage} groups the same 100 families by whether a leak-safe repair unlocked them. Top-level categories are mutually exclusive and reflect the \emph{final} interpretation of each family used in Figure~\ref{fig:audit_triage}.

\begin{table}[h]
\centering
\small
\caption{Final triage of the $n{=}100$ unrecovered audit sample after leak-safe repair attempts. Top-level categories are mutually exclusive and sum to 100. The pessimistic upper bound on composition-related failures treats all 17 leak-safe decomposition repairs as plausibly suspect (residual risk of partial \textsc{Final\_Synthesis}-style leakage even after the symbolic-equivalence filter).}
\label{tab:audit_triage}
\vspace{4pt}
\begin{tabular}{lr}
\toprule
\textbf{Final triage category} & \textbf{Count} \\
\midrule
\textsc{Artifact} (verifier-repaired or leak-safe decomposition-repaired) & 37 \\
\quad verifier-repair fixes (19/25 attempted verifier-artifact families unlocked) & 19 \\
\quad leak-safe decomposition repairs (17/45 attempted families unlocked) & 17 \\
\quad verifier-artifact whose repair was not pipeline-applicable & 1 \\
\midrule
\textsc{Composition-related or unresolved after repair} & 57 \\
\quad rater-confirmed composition failures, kept as-is & 18 \\
\quad still unrecovered after leak-safe decomposition repair & 33 \\
\quad still unrecovered after verifier-repair & 6 \\
\midrule
\textsc{Beyond capability} (kept as-is) & 5 \\
\textsc{Not standalone} (kept as-is) & 1 \\
\midrule
\textbf{Total} & \textbf{100} \\
\midrule
\textit{Pessimistic upper bound on composition-related failures: $57+17=74$.} & \\
\bottomrule
\end{tabular}
\end{table}

\paragraph{Per-category post-repair oracle counts.} Table~\ref{tab:repair_results} reports the fingerprint after the repair attempt for each rater-defined category, on the 100-family audited subset (MilestoneRL-2K step 180, $K{=}8$). The leak-safe decomposition repair enforces condition~(1) of the leak-safe rubric in Appendix~\ref{app:leakage_rubric} (no regenerated milestone is symbolically equivalent to the parent answer) and is therefore restricted to the 45 of 50 decomposition-incomplete families where this filter can be satisfied. The remaining 5 fail condition~(1) and are excluded. Conditions~(2)--(3), which require human or stronger symbolic checks for partial leakage, motivate the pessimistic upper bound rather than the per-category re-run reported here.

\begin{table}[h]
\centering
\small
\caption{\textbf{Oracle re-evaluation after repair on the audited subset.} Rows compare leak-safe decomposition repair, verifier repair, and the rater-confirmed genuine subset on a fresh oracle pass.}
\label{tab:repair_results}
\vspace{4pt}
\setlength{\tabcolsep}{4pt}
\resizebox{\linewidth}{!}{%
\begin{tabular}{lrcccc}
\toprule
\textbf{Source} & \textbf{$n$} & \textsc{Direct} ($C_1$) & \textsc{Roadmap-Needed} & \textsc{Answers-Needed} & \textsc{Unrecovered} \\
\midrule
Leak-safe decomposition repair (45/50) & 45 & 7 & 3 & 7 & 28 \\
Verifier-repair (25/26 attempted) & 25 & 10 & 6 & 3 & 6 \\
Rater-confirmed genuine, re-run fingerprint & 18 & 6 & 3 & 2 & 7 \\
Kept beyond capability (re-run) & 5 & 0 & 1 & 1 & 3 \\
Kept not standalone (re-run) & 1 & 0 & 0 & 1 & 0 \\
\bottomrule
\end{tabular}}

\vspace{2pt}
\parbox{\linewidth}{\footnotesize \emph{Note.} For comparison, the leaky natural-repair version (47 families re-evaluated with no leak-safe filter) gives $D{=}10$, $R{=}31$, $A{=}3$, $U{=}3$. This is reported as a transparency note. The leaky repair is not used in the main paper because every repaired family contains a milestone symbolically equivalent to the parent answer.}
\end{table}

\paragraph{Sampling stochasticity in the re-run.} Table~\ref{tab:repair_results} also exposes the sampling variation inherent to $K{=}8$ rollouts at temperature 1.0. Each row reports a fresh oracle pass on the audited families. The original triage classifies each family from a single 8-rollout pass per condition. \textsc{Unrecovered} in that pass means 0 successes in 8 attempts. A fresh re-run draws 8 new rollouts, and a family with even a small true success rate can produce $\geq 1$ success on the new draw and move out of \textsc{Unrecovered}. The ``Rater-confirmed genuine, re-run fingerprint'' row makes this concrete: of 18 families whose original triage was \textsc{genuine composition} (still \textsc{Unrecovered} under all three probes), 11 produce at least one success on the re-run and only 7 remain \textsc{Unrecovered}. We report both the original triage (used for Section~\ref{sec:repair}) and the re-run fingerprint side by side rather than collapsing them, so the share of the residual that is sampling-stable is visible.

\paragraph{$C_3$ leak residual on the released 354-family set.} The leak-safe repair rule of Appendix~\ref{app:leakage_rubric} is applied as a per-repair check inside the audit. It is not used as a preprocessing filter on the diagnostic-set construction itself. We re-audit the released 354-family set against three leak criteria of increasing strictness: (i)~strict whitespace-normalized equality between an intermediate milestone's gold answer and the parent answer, (ii)~equality after additionally stripping braces, parentheses, and whitespace (trivial-format equivalents), and (iii)~symbolic equivalence accepted by our verifier cascade. Counts are 10 strict, 13 strict-plus-trivial, and 16 under the verifier-accept criterion. We audit the impact on the headline counts using the broader 16-family superset. Per-model $C_3$-only solves on this 16-family leak subset (i.e., families that are \textsc{Answers-Needed} only because a milestone gold answer is equivalent to the parent answer): Qwen-base 0, OutcomeRL-2K 1, MilestoneRL-2K 2, gpt-oss-20b 1, Llama-3.3-70B 1, DeepSeek-V3.1 0, for 5 solves total across the panel. Stripping these 5 solves moves at most 2 families per model from \textsc{Answers-Needed} to \textsc{Unrecovered}, leaves the 33--48\% \textsc{composition-gap} range and the cross-model ordering on Solved-by-any-probe in Table~\ref{tab:same_set_summary} unchanged, and pushes the audit's three-tier reading on \textsc{Unrecovered} only mildly (the upper-tier 74/100 is invariant, and the middle-tier 57/100 increases by at most 2).

\subsection{Audit Provenance, Automated Review, and Raw-Output Retest}
\label{app:audit_rebuttal}

\paragraph{Who labeled the audit.}
Two of the authors labeled the 100 unrecovered MilestoneRL-2K families of Table~\ref{tab:audit_main}. Each rater labeled every family independently with the rubric above. The raters agreed on 67 families, and the fixed priority rule resolved the other 33, 14 to \textsc{Verifier\_Artifact}, 3 to \textsc{Beyond\_Capability}, and 16 to \textsc{Decomposition\_Incomplete}. Both raters' labels, their one-line reasons, and the resolution of every disagreement are released. Both raters are human annotators from the author team, and they knew that every sampled family was unrecovered. A separate human validation pass on 50 of the families gives the agreement statistics in Appendix~\ref{app:audit_details}. The raw-output retest below re-reads the 18 families that both raters labeled \textsc{Genuine\_Composition}, and 5 of them turn out to carry a defective reference answer (Table~\ref{tab:defective_refs}).

\paragraph{Automated rubric review of all six models.}
To measure the grading share of \textsc{composition-gap} for every model, we reviewed each model's \textsc{composition-gap} families with the same five-category rubric. The review covers the families in the category under the 4K milestone test, the snapshot available when the review was run, which gives 906 model and family pairs (Table~\ref{tab:stage0_budget}). The labels come from a Codex CLI agent running GPT-5.5 at high reasoning effort, given the rubric document that the artifact releases. The agent labeled each of the 253 distinct families once and copied the label to every model that has the family in the category. It marked 61 families as verifier or format errors, 38 of them through a written judgment about the family and 23 through rules on the reference-answer text, such as garbled or truncated \LaTeX{}, a problem that depends on a figure, or a multi-part question with a single-value reference answer. It labeled 186 families \textsc{Decomposition\_Incomplete}, all with one shared justification, 2 families \textsc{Not\_Standalone}, and 4 families \textsc{Genuine\_Composition}, each of them a family where a milestone answer already equals the reference answer. Because the decomposition and composition labels share one justification, we use only the review's verifier and format axis.

Table~\ref{tab:rubric_review} gives the result. The review flags 22.2--29.1\% of each model's \textsc{composition-gap} families as verifier or format errors. Removing them lowers the category from 34--49\% to 25--37\% of the 354 families on the 4K snapshot. Mapping each family's label onto the 16K sets of the main text lowers it from 33--48\% to 24--37\%, with 7 of the 907 model and family pairs unlabeled and kept in the category. The ranking of the six models changes only between models within one percentage point of each other.

\begin{table}[h]
\centering
\footnotesize
\caption{Automated rubric review of each model's \textsc{composition-gap} families. Flagged families are those labeled verifier or format errors. The last four columns give the category's share of the 354 families before and after removing the flagged families, on the 4K snapshot the review labeled and on the 16K sets of the main text with each family's label carried over.}
\label{tab:rubric_review}
\vspace{4pt}
\setlength{\tabcolsep}{4pt}
\begin{tabular}{lrrrrrrr}
\toprule
 & \multicolumn{3}{c}{\textbf{4K families labeled}} & \multicolumn{2}{c}{\textbf{Share, 4K}} & \multicolumn{2}{c}{\textbf{Share, 16K}} \\
\cmidrule(lr){2-4} \cmidrule(lr){5-6} \cmidrule(lr){7-8}
\textbf{Model} & \textbf{Families} & \textbf{Flagged} & \textbf{Flagged \%} & \textbf{Before} & \textbf{After} & \textbf{Before} & \textbf{After} \\
\midrule
Qwen-base & 173 & 43 & 24.9 & 48.9 & 36.7 & 48.3 & 36.7 \\
OutcomeRL-2K & 155 & 39 & 25.2 & 43.8 & 32.8 & 43.8 & 32.5 \\
MilestoneRL-2K & 167 & 37 & 22.2 & 47.2 & 36.7 & 47.2 & 36.4 \\
gpt-oss-20b & 120 & 33 & 27.5 & 33.9 & 24.6 & 33.3 & 24.3 \\
Llama-3.3-70B-Instruct & 164 & 38 & 23.2 & 46.3 & 35.6 & 48.3 & 37.0 \\
DeepSeek-V3.1 & 127 & 37 & 29.1 & 35.9 & 25.4 & 35.3 & 25.4 \\
\bottomrule
\end{tabular}
\end{table}

The review and the author audit flag different families. On the 74 families that both cover, they give the same verifier or format verdict for 51 (69\%, Cohen's $\kappa{=}0.14$), and the review flags 6 of the 20 families that the authors labeled \textsc{Verifier\_Artifact}. The 22--29\% is therefore one automated estimate of the grading share. The frontier-judge adjudication (Appendix~\ref{app:frontier_judge}) and the retest below measure the same layer by other means, and both find grading and reference errors common.

\paragraph{Raw-output retest of the 18 genuine-composition families.}
The strongest direct evidence comes from one family. On ff548f26, Qwen3-8B with thinking disabled solves both milestones alone in 8 of 8 attempts, receives both correct milestone answers, and still fails all eight attempts at the full problem, while gpt-oss-20b solves it from the same roadmap and answers in 6 of 8. The retest covers the 18 families that both raters labeled \textsc{Genuine\_Composition}. They come from MilestoneRL-2K, and neither MilestoneRL-2K nor Qwen-base can be sampled any more, because the serving platform retired their base checkpoint. We therefore used Qwen3-8B with thinking disabled, the post-trained model of the same size and family, as a stand-in 8B student, next to gpt-oss-20b. Each model received the $C_3$-gold prompt of every family, the roadmap plus every gold milestone answer, 8 times at \texttt{max\_tokens=16384}. The prompt text is the panel's $C_3$-gold prompt, and the retest omitted the panel's system prompt and sampled at temperature 0.7 in place of 1.0. We saved every response and graded it by hand against a verified answer, counting an attempt as correct when its boxed answers state the correct parent answer.

Table~\ref{tab:e1_retest} gives the counts. With every milestone answer in hand, Qwen3-8B (thinking disabled) answers correctly in at least one of eight attempts on 16 of the 18 families, and gpt-oss-20b does so on all 18. Two families separate the models. On ff548f26, described above, Qwen3-8B reports an area of 5 or 65/4 in every attempt, while the true area is 19/4. On 0ad0ce27, Qwen3-8B is correct in 1 of 8 attempts against 8 of 8 for gpt-oss-20b, although it solves both milestones alone in 8 of 8.

The other family where Qwen3-8B fails all eight attempts, a0759695, has wrong gold milestone answers. Its roadmap factors the parabola as $(x-1)(x+a+2)$, which treats $x{=}1$ as a root although the parabola passes through $(1,-1)$, and two later milestone answers, the root interval and the scaling coefficient, inherit the error. Qwen3-8B follows the given answers and reports $\pi/30$, $0$, or an expression in $a$. gpt-oss-20b states the contradiction in several attempts, re-derives the volume $\frac{\pi}{30}(a^2+4a+8)^{5/2}$, and reaches the true minimum $\frac{16\pi}{15}$ in 7 of 8 attempts. On ad098d0c, which the verifier scores 0 of 8, every Qwen3-8B attempt boxes the correct maximum $\frac{3\sqrt{2}}{4}$ and then boxes the correct set of maximizers, and the cascade reads only the last box. On most of the remaining families both models are right, and the obstacle is the reference answer or the answer format listed in the table.

These 18 families were selected by the author audit, so the counts describe the audited subset and serve as evidence that the failure exists, while the per-model grading share comes from the review above.

\begin{table}[h]
\centering
\footnotesize
\caption{Raw-output retest of the 18 families that both raters labeled \textsc{Genuine\_Composition}. Each model receives the $C_3$-gold prompt 8 times ($K{=}8$, \texttt{max\_tokens=16384}, temperature 0.7). Cells count attempts whose boxed answers state the correct parent answer, graded by hand against a verified answer. On 29eb7827, all 16 attempts state the correct order in words and 2 box it.}
\label{tab:e1_retest}
\vspace{4pt}
\begin{tabular}{llrr}
\toprule
\textbf{Family} & \textbf{Reference answer or format issue} & \textbf{Qwen3-8B} & \textbf{gpt-oss-20b} \\
\midrule
ff548f26 & none & 0 & 6 \\
0ad0ce27 & none & 1 & 8 \\
a0759695 & wrong reference and wrong milestone answers & 0 & 7 \\
ad098d0c & two-part question, the cascade reads the last box & 8 & 8 \\
\midrule
713ff8c1 & wrong reference answer & 8 & 6 \\
a3a636a8 & malformed reference, a milestone answer gives the answer & 8 & 8 \\
330cbc5a & malformed reference, variable name missing & 7 & 8 \\
c52e7627 & reference lists one of two solutions & 7 & 7 \\
5ce4a16b & two-part reference answer & 8 & 8 \\
20dfe8b1 & letter reference, answer given as its value & 6 & 8 \\
60f12f06 & letter reference & 7 & 7 \\
29eb7827 & asks for an order, reference is an expression & 1 & 1 \\
40a06b35 & answer is a condition stated in words & 8 & 5 \\
cb760e44 & hyperbolic-function notation & 8 & 8 \\
2e647bfc & answer boxed as an equation & 8 & 7 \\
723ca360 & extra text in the boxed answer & 6 & 8 \\
eeab310c & none & 7 & 7 \\
a0146340 & none & 8 & 8 \\
\bottomrule
\end{tabular}
\end{table}

\paragraph{Defective reference answers.}
Table~\ref{tab:defective_refs} lists the defective reference answers that the retest and the frontier-judge adjudication exposed in the 354-family set, each checked by hand. Five of them belong to the 18 families that both raters labeled \textsc{Genuine\_Composition}. The release keeps the original reference answers, which all counts in this paper use, and adds each corrected answer in a separate field together with the evidence for the change. The three wrong milestone answers of a0759695 are marked in the same way. A corrected answer can change the category of its family for a model, so these ten families move any model's \textsc{composition-gap} share by at most 2.8 percentage points.

\begin{table}[h]
\centering
\footnotesize
\caption{Defective reference answers found in the 354-family set. The first five families belong to the 18 audited genuine-composition families.}
\label{tab:defective_refs}
\vspace{4pt}
\begin{tabular}{lll}
\toprule
\textbf{Family} & \textbf{Reference answer in the set} & \textbf{Defect and verified answer} \\
\midrule
713ff8c1 & \texttt{48884} & wrong, largest palindrome is $49894 = 101 \cdot 494$ \\
a0759695 & \texttt{\textbackslash frac\{56\textbackslash pi\}\{15\}} & wrong, minimum volume is $16\pi/15$ \\
a3a636a8 & \texttt{\textbackslash leqslant-2} & variable missing, answer is $a \leq -2$ \\
330cbc5a & \texttt{=2,b=\textbackslash frac\{1\}\{4\}} & variable missing, answer is $a{=}2$, $b{=}1/4$ \\
c52e7627 & \texttt{x\_\{1\}=\textbackslash frac\{15\}\{2\},y\_\{1\}=...} & second solution $(3/2, 1/2)$ missing \\
b36a1161 & \texttt{\textbackslash 1} & garbled, answer is $x = \pm 1$ \\
41348e33 & \texttt{x\_1=x\_2=} & truncated, answer is $x_1 = x_2$, a nonzero integer \\
860813bf & long list ending in \texttt{n=1,} & truncated, answer is $n \in \{2, 3\}$ \\
e24a0cbb & \texttt{0\textbackslash leqslant|+b|\textbackslash leqslant2} & garbled, the term $az$ is missing \\
cdfc31d2 & \texttt{15to25} & words run together \\
\bottomrule
\end{tabular}
\end{table}

\section{Leak-Safe Repair Rubric and Examples}
\label{app:leakage_rubric}

This appendix specifies the leak-safe repair rule used in the audit (Appendix~\ref{app:audit_details}), with worked examples. The natural repair protocol initially tested generates a regenerated milestone set that ends with a \textsc{Final\_Synthesis} milestone whose gold answer matches the parent answer. We verified by symbolic string normalization that all 47 decomposition-repaired families have at least one milestone whose gold answer is equivalent to the parent answer. Under condition $C_3$, the student is then handed that answer verbatim, so $C_3$ success on a decomposition-repaired family is not evidence that the family was solved through composition.

\paragraph{Leak-safe rule.} A repaired family is \emph{leak-safe} if and only if:
\begin{enumerate}[nosep,leftmargin=*]
  \item no milestone gold answer is symbolically equivalent to the parent gold answer,
  \item no milestone gold answer is a trivial transformation of the parent answer (e.g., unit rewriting, sign flip, case split recombination),
  \item no milestone collapses the parent solve into a single last formatting step (e.g., ``given $s^2 = 884 - 440\sqrt{3}$, state the area'').
\end{enumerate}
Condition~(1) is mechanically checkable via the verifier cascade. Conditions~(2) and~(3) are conservative checks that require a small human pass, a stronger symbolic normalizer, or a teacher-side constraint at generation time (the last milestone must leave the student with at least one non-trivial composition step before the parent answer can be written down).

\subsection*{Repair leakage examples}

\paragraph{Disallowed under strict rule: leaky decomposition repair.}
\textbf{Example} (audit category: \textsc{Decomposition\_Incomplete}).\\
\textit{Parent (truncated):} ``Josef returns from a trip by train then bike; the entire journey takes 1 hour 30 minutes over 60 km; the train traveled at 50 km/h; the bike speed $v$ km/h is a natural number, as is the distance in $v$ km. Determine how long Josef rode his bike.''\\
\textit{Parent answer:} \verb|30,36,60|

\textit{Repaired milestones:}
\begin{itemize}[nosep,leftmargin=*]
  \item M1 (\textsc{Model}): Let $v$ be the bike distance in km and $r$ the bike speed in km/h. Derive the travel-time equation. \quad Answer: \verb|(60-v)/50 + v/r = 3/2|
  \item M2 (\textsc{Key\_Move}): Rewrite the equation as a product of two linear factors. \quad Answer: \verb|(50-r)(v+15)=750|
  \item M3 (\textsc{Compute}): Enumerate natural-number pairs $(v,r)$. \quad Answer: \verb|(10,20),(35,15),(60,10)|
  \item M4 (\textsc{Final\_Synthesis}): State the final bike-time answers. \quad Answer: \verb|30,36,60| \textbf{[LEAK: $\equiv$ parent answer]}
\end{itemize}
\textit{Post-repair oracle (Mile-2k step 180, $K{=}8$):} $C_1{=}0$, $C_2{=}0$, $C_3{=}2$.
The $C_3$ success is not informative because the student is handed the final answer verbatim in M4.

\paragraph{Allowed under strict rule: leak-safe verifier repair (no new milestones).}
\textbf{Example} (audit category: \textsc{Verifier\_Artifact}).\\
\textit{Parent:} ``A point moves along a straight line with constant acceleration. Find the law of motion of the point.''\\
\textit{Parent answer:} \verb|s(t) = (1/2)at^2 + v_0*t + s_0|

\textit{Milestones (unchanged):}
\begin{itemize}[nosep,leftmargin=*]
  \item M1 (\textsc{Model}): Write the ODE for velocity. \quad Answer: \verb|dv/dt = a|
  \item M2 (\textsc{Key\_Move}): Solve for $v(t)$ with $v(0){=}v_0$. \quad Answer: \verb|at + v_0|
\end{itemize}
\textit{Repair action:} the original canonical answer was a corrupted LaTeX string. The verifier rule is tightened to accept the normalized form. No new milestones are introduced, so the parent answer does not appear in any milestone.\\
\textit{Post-repair oracle:} $C_1{=}4$, $C_2{=}2$, $C_3{=}7$.

\paragraph{Families labeled genuine composition (no repair attempted).} For 18 of the 100 audited families both raters labeled \textsc{Genuine\_Composition}, and the student fails all three parent probes. A raw-output retest of these 18 families (Appendix~\ref{app:audit_rebuttal}) finds a defective reference answer in 5 of them and wrong gold milestone answers in 1, and it shows a clean failure to combine given steps on \texttt{ff548f26} for Qwen3-8B (thinking disabled).

Table~\ref{tab:repair_results} reports the leak-safe re-run on the 45 decomposition-incomplete families that satisfy condition~(1).

\section{Teacher Prompt (Excerpt)}
\label{app:teacher_prompt}

This appendix excerpts the teacher-side compilation prompt. The full prompt is released with the artifact. The teacher model (GPT-5.4 Thinking) is instructed to produce typed milestone decompositions, with three key rules governing standalone solvability, coverage, and schema enforcement:

\begin{quote}
\small
\textit{``You are a Mathematics Mentor. Your ONLY job is to output a milestone plan in valid JSON.}

\textit{CRITICAL STANDALONE RULE: Each milestone must be an individually solvable mini-problem. A solver must be able to answer a milestone using ONLY that milestone's own description (plus the original problem context). You MUST NOT rely on any unstated lemma, theorem, definition, or earlier milestone text.}

\textit{COVERAGE RULE: The milestone set must cover the entire solution path so that following milestones sequentially leaves no hidden hard steps. Avoid magic-jump wording unless the milestone also states an explicit, checkable target.}

\textit{Each milestone specifies: type (KEY\_MOVE, MODEL, COMPUTE, NORMALIZE, INTEGRATE, LEMMA, WARMUP, SANITY), answer\_schema (NUMBER, EXPR, EQUATION, INEQUALITY, SET, INTERVAL, TUPLE\_LIST, MATRIX\_VECTOR, MCQ\_STRING), description, and canonical\_answer.''}
\end{quote}

The teacher chooses the answer schema \emph{before} writing the milestone description, ensuring the output format is verifiable. Milestones that fail schema validation are rejected and regenerated.

\section{Verifier Cascade: Extended Characterization}
\label{app:verifier_ext}

This appendix supports the verifier-design choice from Section~\ref{sec:setup} with a 400-response paired audit comparing the strict symbolic cascade with an 8B LLM judge, plus per-schema and false-negative characterization. The paired audit measures agreement relative to the gold answers, and Appendix~\ref{app:frontier_judge} adds a grader-blind adjudication of disagreements against the responses themselves. The strict cascade has zero observed false positives on the 400 paired sample (95\% upper bound 0.75\% by the rule of three). It errs in the other direction. A grader-blind adjudication of frontier-judge disagreements (Appendix~\ref{app:frontier_judge}) estimates that it rejects a correct answer or grades against a defective reference in about 8.7\% of graded responses, which is 10.5\% of the responses it rejects. The 8B judge accepts 14.75\% of paired responses that the cascade rejects.

\paragraph{Paired audit setup.} We assembled 400 student responses spanning both in-distribution NuminaMath milestone prompts and out-of-distribution AIME responses, covering all nine answer schemas. Each response was graded by two procedures: the strict symbolic cascade (boxed extraction $\to$ schema-aware parse $\to$ symbolic-algebra equivalence $\to$ \texttt{math-verify}~\citep{mathverify2025} symbolic check, abstaining if any stage fails), and an 8B LLM-as-judge (Qwen3-8B-Base in no-think mode with a boxed-extraction prompt). The paired comparison is against the canonical (i.e., gold) answer.

\paragraph{Agreement and disagreement structure.} Of 400 paired grades: 217 both-ACCEPT (54\%), 124 both-NOT\_ACCEPT (31\%), 59 disagreements (15\%). \emph{Every disagreement is strict-NOT\_ACCEPT / LLM-ACCEPT.} There are zero strict-ACCEPT / LLM-NOT\_ACCEPT cases in this sample. Under the assumption that the gold answer is the target (i.e., a response is ``correct'' iff symbolically equivalent to the canonical), the 59 disagreements are an upper bound on the judge's over-acceptance. The frontier-judge comparison shows that most such disagreements are cascade errors, and that a judge rejects a cascade-accepted response only rarely (0 of 400 here, 1 of 997 for gpt-5.5). We observe zero strict-cascade false positives in 400 paired examples. By the rule-of-three, the 95\% upper confidence bound on the strict-cascade false-positive rate is 3/400 = 0.75\%.

\begin{table}[h]
\centering
\small
\caption{Per-schema breakdown of the 59 strict-NOT\_ACCEPT / LLM-ACCEPT disagreements. \textsc{Expr} (formulas) and \textsc{Number} dominate. For the frontier judge, disagreements of this kind are mostly correct answers that the cascade cannot parse (Appendix~\ref{app:frontier_judge}).}
\label{tab:verifier_schema}
\vspace{4pt}
\begin{tabular}{lrrr}
\toprule
\textbf{Schema of expected answer} & \textbf{Disagreements} & \textbf{Share} & \textbf{LLM over-accept rate in schema} \\
\midrule
\textsc{Expr} (free-form formula) & 34 & 58\% & 34/156 = 22\% \\
\textsc{Number} & 24 & 41\% & 24/224 = 11\% \\
\textsc{Interval}/\textsc{Tuple} & 1 & 2\% & 1/12 = 8\% \\
\textsc{Equation}, \textsc{Set}, other & 0 & 0\% & 0/8 = 0\% \\
\bottomrule
\end{tabular}
\end{table}

\paragraph{False negatives of the strict cascade.} A strict false negative is a correct response that the cascade rejects. The frontier-judge adjudication measures this directly (Appendix~\ref{app:frontier_judge}). About 68\% of the responses that gpt-5.5 accepts and the cascade rejects are correct answers the cascade cannot parse or match, or cases graded against a defective reference answer. This puts the cascade's grading errors at about 8.7\% of graded responses, concentrated on NuminaMath parent answers with compound or unusual forms, and at 0.7\% on AIME integer answers. The raw-output retest (Appendix~\ref{app:audit_rebuttal}) finds the same kinds of errors inside the audited \textsc{composition-gap} families.

\paragraph{Interpretation.} The cascade and an LLM judge err in opposite directions. The cascade rejects some correct answers in unusual forms, and we measure and remove this layer inside each model's \textsc{composition-gap} (Appendix~\ref{app:audit_rebuttal}). An LLM judge credits unfinished work, which raises the success rate of any condition that produces long partial solutions and would blur the corruption controls of the specificity result. We therefore keep the deterministic cascade and report its strictness.

\subsection{Frontier-Judge Comparison}
\label{app:frontier_judge}

\paragraph{Setup.}
We repeated the paired audit with a frontier judge, gpt-5.5 from OpenAI, using the same judge prompt as the 8B judge above. The prompt gives the verifier system prompt, the reference answer, the grading note, and the response, and asks for a JSON verdict at temperature 0. The sample has 997 responses after 3 API errors are removed. Of these, 698 are parent attempts from the format audit (Appendix~\ref{app:decoding_budget}) by Qwen-base, MilestoneRL-2K, Llama-3.3-70B-Instruct, and DeepSeek-V3.1 under $C_1$ and $C_3$-gold at \texttt{max\_tokens=4096}, and 299 are AIME 2024/25 attempts by RL checkpoints of the Qwen lineage. Both graders read the last 12{,}000 characters of each response, and the cascade runs without any LLM fallback.

\paragraph{Paired grades.}
The frontier judge shows the same one-sided pattern as the 8B judge (Table~\ref{tab:frontier_paired}). It accepts 127 responses (12.7\%) that the cascade rejects and rejects 1 response (0.1\%) that the cascade accepts. The disagreements fall almost entirely on NuminaMath parent problems, whose reference answers take many forms (125 of 698, 17.9\%), and are rare on AIME, whose answers are integers (2 of 299, 0.7\%).

\begin{table}[h]
\centering
\footnotesize
\caption{Paired grades of the strict cascade and two LLM judges against the reference answer. The first row is the 400-response audit above. The other rows use gpt-5.5 on 997 responses, in total and by source.}
\label{tab:frontier_paired}
\vspace{4pt}
\setlength{\tabcolsep}{4pt}
\begin{tabular}{lrrrrr}
\toprule
 & & \textbf{Both} & \textbf{Both} & \textbf{Judge accepts,} & \textbf{Cascade accepts,} \\
\textbf{Judge and responses} & $n$ & \textbf{accept} & \textbf{reject} & \textbf{cascade rejects} & \textbf{judge rejects} \\
\midrule
8B judge, 400-response audit & 400 & 217 & 124 & 59 (14.75\%) & 0 (0.0\%) \\
gpt-5.5, all responses & 997 & 167 & 702 & 127 (12.7\%) & 1 (0.1\%) \\
gpt-5.5, NuminaMath parents & 698 & 112 & 460 & 125 (17.9\%) & 1 (0.1\%) \\
gpt-5.5, AIME 2024/25 & 299 & 55 & 242 & 2 (0.7\%) & 0 (0.0\%) \\
\bottomrule
\end{tabular}
\end{table}

\paragraph{Grader-blind adjudication.}
To find out which grader is right when they disagree, we drew 40 of the 127 disagreements, stratified by the reason the cascade gave for rejecting the response. An author adjudicated each case while seeing only the end of the response, the reference answer, and the grading note, and never either grader's verdict. Six responses ended before any answer could be judged, which leaves 34 adjudicated cases. Of these, 18 are correct answers that the cascade could not parse or match, 6 are graded against a defective reference answer, and 10 are genuine over-acceptance by the judge. The release includes the 40 sampled cases and the counts per stratum.

The verdict follows the cascade's failure mode almost exactly (Table~\ref{tab:frontier_adjudication}). When the cascade reports a parse error, 21 of the 22 adjudicated cases are grading errors on the cascade side, for example the same value written in two algebraic forms or a reference answer with a multiple-choice letter attached. When the response contains no boxed answer, 8 of the 9 cases are judge errors, and all 8 are the same failure, a response that stops mid-reasoning without a final answer and that the judge credits for its reasoning.

\begin{table}[h]
\centering
\footnotesize
\caption{Grader-blind adjudication of judge-accept, cascade-reject disagreements, by the cascade's failure mode. The cascade side counts correct answers the cascade could not match and cases graded against a defective reference answer. The last row weights each failure mode by its share of the 127 disagreements.}
\label{tab:frontier_adjudication}
\vspace{4pt}
\setlength{\tabcolsep}{4pt}
\begin{tabular}{lrrrrr}
\toprule
\textbf{Cascade failure mode} & \textbf{Disagreements} & \textbf{Sampled} & \textbf{Adjudicated} & \textbf{Cascade side} & \textbf{Judge side} \\
\midrule
Parse error & 75 & 24 & 22 & 21 & 1 \\
No boxed answer & 35 & 11 & 9 & 1 & 8 \\
Not equivalent & 17 & 5 & 3 & 2 & 1 \\
\midrule
All modes & 127 & 40 & 34 & 24 & 10 \\
Weighted to all 127 & & & & 86.8 (68\%) & 40.2 (32\%) \\
\bottomrule
\end{tabular}
\end{table}

\paragraph{Weighted split and the cascade's strictness.}
Weighting each failure mode by its share of the 127 disagreements (75 parse errors, 35 responses without a boxed answer, and 17 non-equivalent answers) attributes about 87 disagreements (68\%) to cascade strictness or defective reference answers and about 40 (32\%) to the judge. On this sample, the cascade therefore rejects a correct answer or grades against a defective reference in about 8.7\% of the 997 graded responses, which is 10.5\% of the 829 responses it rejects. The judge's own over-acceptance covers about 4.0\% of graded responses.

\paragraph{Reading.}
The two graders err in opposite directions and in different places. The cascade rejects correct answers written in unusual forms and follows defective reference answers, which lowers measured recovery and moves some families into \textsc{Unrecovered}. We measure this layer inside each model's \textsc{composition-gap} (Appendix~\ref{app:audit_rebuttal}) and remove it before reading the category. The judge credits unfinished work, which raises the success rate of any condition that produces long partial solutions, and the controls of Section~\ref{sec:oracle_specificity} depend on telling exactly such conditions apart. We therefore keep the deterministic cascade, which gives the same verdict on every rerun and can be inspected rule by rule, and we report its strictness next to the results it grades.

\section{Answer Schema Definitions}
\label{app:schemas}

This appendix lists the distribution of the nine answer schemas used by the teacher pipeline. Each schema has a deterministic parser and a schema-specific equivalence check.

\begin{table}[h]
\centering
\small
\begin{tabular}{llr}
\toprule
\textbf{Schema} & \textbf{Example} & \textbf{\%} \\
\midrule
EXPR & $\boxed{\frac{a^2}{8}}$ & 44.6 \\
EQUATION & $\boxed{y = 13 - 2x}$ & 19.2 \\
NUMBER & $\boxed{42}$ & 15.7 \\
INEQUALITY & $\boxed{n \geq 7}$ & 6.1 \\
SET & $\boxed{\{1, 2, 3\}}$ & 5.7 \\
TUPLE\_LIST & $\boxed{(1,2), (3,4)}$ & 4.9 \\
MCQ\_STRING & $\boxed{D}$ & 2.6 \\
INTERVAL & $\boxed{[0, \pi/2]}$ & 1.0 \\
MATRIX\_VECTOR & $\boxed{\begin{pmatrix} 1 \\ 0 \end{pmatrix}}$ & 0.2 \\
\bottomrule
\end{tabular}
\caption{Answer schema distribution in the pipeline (all milestones, $n{=}4654$).}
\end{table}

\section{Author Contributions}
\label{app:contributions}
Zhuohan Wang conceived and led the project, designed OracleLadder, ran the experiments, and wrote the paper. Haoran Ma and Tianyu Wu annotated the audit data. Yuanlin Duan helped develop the idea and contributed to the writing. Zichun Liao contributed to discussions of the ideas. Jieming Yu contributed to the writing.

\end{document}